\documentclass{article}

\usepackage{arxiv}

\usepackage[utf8]{inputenc} 
\usepackage[T1]{fontenc}    
\usepackage{hyperref}       
\usepackage{url}            
\usepackage{booktabs}       
\usepackage{amsfonts}       
\usepackage{nicefrac}       
\usepackage{graphicx}
\usepackage{natbib}
\usepackage{doi}
\usepackage{bm}
\usepackage{makecell}
\usepackage{multirow}
\usepackage[textsize=tiny]{todonotes}
\usepackage{xcolor, soul}
\usepackage{threeparttable}
\usepackage{adjustbox}
\usepackage{siunitx} 
\usepackage{textcomp}
\usepackage{gensymb}
\usepackage{tikz}
\usepackage{amsmath}
\usepackage{cases}

\usepackage{tabularx,booktabs}
\newcolumntype{Y}{>{\centering\arraybackslash}X}
\usepackage{tabularray}

\usepackage{setspace}
\usepackage[c1]{optidef}
\usepackage{silence}
\title{
The interplay of fatigue dynamics and task achievement using optimal control predictive simulation.}

\author{ \href{https://orcid.org/0000-0002-9335-630X}{\includegraphics[scale=0.06]{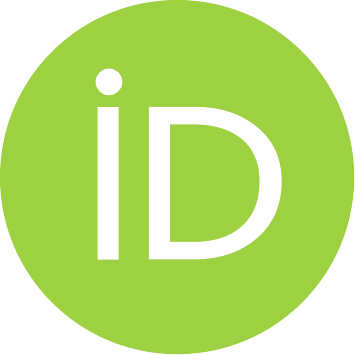}\hspace{1mm}P. Puchaud\dag\textasteriskcentered}, \href{https://orcid.org/0000-0002-5031-1048}{\includegraphics[scale=0.06]{orcid.pdf}\hspace{1mm}B. Michaud\dag}, \href{https://orcid.org/0000-0002-9335-630X}{\includegraphics[scale=0.06]{orcid.pdf}\hspace{1mm}M. Begon} \\
	Laboratoire de Simulation et Modélisation du Mouvement \\
    École de kinésiologie et des Sciences de l’Activité Physique \\
    Université de Montréal, Laval, Canada\\
    Centre de réadaptation Marie-Enfant \\
    CHU Sainte-Justine, Montréal, Canada \\
	\texttt{pierre.puchaud@umontreal.ca} \\
    \textasteriskcentered~Corresponding author \\
    \dag~Authors equally contributed
}

\date{13th June 2023}

\renewcommand{\shorttitle}{The interplay of fatigue dynamics and task achievement using optimal control.}

\DeclareFontFamily{U}{mathb}{\hyphenchar\font45}
\DeclareFontShape{U}{mathb}{m}{n}{
      <5> <6> <7> <8> <9> <10> gen * mathb
      <10.95> mathb10 <12> <14.4> <17.28> <20.74> <24.88> mathb12
      }{}
\DeclareSymbolFont{mathb}{U}{mathb}{m}{n}
\DeclareFontSubstitution{U}{mathb}{m}{n}

\let\dot\relax
\DeclareMathAccent{\dot}{0}{mathb}{"39}
\let\ddot\relax
\DeclareMathAccent{\ddot}{0}{mathb}{"3A}
\let\dddot\relax
\DeclareMathAccent{\dddot}{0}{mathb}{"3B}
\let\ddddot\relax
\DeclareMathAccent{\ddddot}{0}{mathb}{"3C}

\newcommand{\qddot}{\bm{\ddot{q}}}
\newcommand{\qdot}{\bm{\dot{q}}}
\newcommand{\q}{\bm{q}}

\newcommand{\btau}{\bm{\tau}}
\newcommand{\controls}{\bm{\mathrm{u}}}
\newcommand{\states}{\bm{\mathrm{x}}}

\newcommand{\statesdot}{\bm{\dot{\mathrm{x}}}}

\DeclareFontFamily{U}{mathc}{}
\DeclareFontShape{U}{mathc}{m}{it}%
{<->s*[1.03] mathc10}{}
\DeclareMathAlphabet{\mathscr}{U}{mathc}{m}{it}

\newcommand{\tl}{T\!L}

\newcommand{\mstate}{\mathrm{m}}
\newcommand{\actives}{a}
\newcommand{\rest}{r}
\newcommand{\fatigue}{f}

\newcommand{\ma}{\mstate_\actives}
\newcommand{\mf}{\mstate_\fatigue}
\newcommand{\mr}{\mstate_\rest}
\newcommand{\bma}{\bm{\mstate}_\actives}
\newcommand{\bmf}{\bm{\mstate}_\fatigue}
\newcommand{\bmr}{\bm{\mstate}_\rest}
\newcommand{\bmdot}{\dot{\bm{\mstate}}}
\newcommand{\mdot}{\dot{\mstate}}
\newcommand{\madot}{\mdot_\actives}
\newcommand{\mfdot}{\mdot_\fatigue}
\newcommand{\mrdot}{\mdot_\rest}

\hypersetup{
pdftitle={The interplay of fatigue dynamics and task achievement using optimal control predictive simulation},
pdfauthor={Pierre Puchaud, Benjamin Michaud, Mickael Begon},
}

\begin{document}
\maketitle

\begin{abstract}
Predictive simulation of human motion could provide insight into optimal techniques. In repetitive or long-duration tasks, these simulations must predict fatigue-induced adaptation.  
However, most studies minimize cost function terms related to actuator activations, assuming it minimizes fatigue.  
An additional modeling layer is needed to consider the previous use of muscles to reveal adaptive strategies to the decreased force production capability. 
Here, we propose interfacing Xia's three-compartment fatigue dynamics model with rigid-body dynamics. A stabilization invariant was added to Xia’s model. We simulated the maximum repetition of dumbbell biceps curls as an optimal control problem (OCP) using direct multiple shooting. 
We explored three cost functions (minimizing minimum torque, fatigue, or both) and two OCP formulations (full-horizon and sliding-horizon approaches).  
We found that Xia's model modified with the stabilization invariant (coefficients $S=\{10,5\}$) was adapted to direct multiple shooting. 
Sliding-horizon OCPs achieved 20 to 21 repetitions. The kinematic strategy slowly deviated from a plausible dumbbell lifting task to a swinging strategy as fatigue onset increasingly compromised the ability to keep the arm vertical. 
In full-horizon OCPs, the latter  kinematic strategy was used over the whole motion, resulting in 32 repetitions. 
We showed that sliding-horizon OCPs revealed a reactive strategy to fatigue when only torque was included in the cost function, whereas an anticipatory strategy was revealed when the fatigue term was included in the cost function. Overall, the proposed approach has the potential to be a valuable tool in optimizing performance and helping reduce fatigue-related injuries in a variety of fields.
\end{abstract}


\section{Introduction}

Predictive simulation of human motion provides insight into the optimal motions to perform a given task such as gait or running \citep{Nitschke2020EfficientDynamics}, acrobatics \citep{Charbonneau2022OptimalSelf-collision}, jumping \citep{Porsa2016DirectOpenSim}, reaching \citep{Maas2013BiomechanicalMotion:}. 
It often relies on optimal control theory.
Optimal control problems (OCPs) provide controls and states that minimize an objective function to accomplish the given task. 
Objective functions include terms such as joint torques \citep{Felis2016SynthesisMethods, Febrer-Nafria2020ComparisonModel}, muscle forces \citep{Ezati2020ComparisonModel}, muscle activations or metabolic cost \citep{Falisse2019, Dembia2019}.
While some authors claimed that minimizing muscle activations is equivalent to minimizing ``fatigue''  \citep{Rasmussen2001MuscleStudy, Halloran2010ConcurrentLoading}, this approach overlooks the long-term accumulation of fatigue in the actuators themselves.
Neuro-muscular fatigue has been defined as a gradual decrease in the maximal force or power that muscles can
produce, during a sustained physical activity \citep{Enoka2008MuscleFunction}. 
There is no single cause of neuro-muscular fatigue, but as computational biomechanists, we commonly model peripheral fatigue as its main mechanism. Peripheral fatigue involves the post-synaptic physiological processes of the motor unit. Mitigating peripheral fatigue necessitates a forward-thinking, middle-term strategy that modifies the entire motion.
A valuable advancement for predictive simulations would be to embed an additional modeling layer for fatigue dynamics within the actuation of models, enabling the system to adjust the global motion by explicitly minimizing a fatigue-related decision variable.

Fatigue dynamics models have been designed to study muscle fatigue at muscle fiber scale \citep{Rockenfeller2020ExhaustionModel} or at the whole muscle scale \citep{Ding2000AMuscles}. These models include several decision variables per muscle (9 and 5, respectively), which would lead to a large number of variable decisions if applied to a full-body musculoskeletal model, making them impractical for musculoskeletal predictive simulation. The theoretical fatigue model of Xia \citep{Xia2008}, also known as the three-compartment model ($3_{CC}$), enables to predict the decay and recovery of the actuator over time. This model includes active ($\ma$), fatigued ($\mf$), and resting ($\mr$) states with a fatigue feedback controller of activation–deactivation added to the original model \citep{Liu2002ARecovery}.
Consistent with muscle physiology, this phenomenological model can predict peripheral fatigue during various tasks.
This model is more generic than empirical ones. 
It can be applied to any actuator torque-driven or muscle-driven simulations, making it flexible for biomechanical predictive simulations.
This model has been used for analyzing and modeling experimental data such as joint-specific maximum endurance times \citep{Frey-Law2012ATasks}, intermittent contractions \citep{Looft2018} or intensity vs endurance time curves \citep{Rakshit2020ModellingJoint}.
It has recently been used to build a criterion to maximize the endurance time in box-carrying tasks when analyzing the static posture \citep{Barman2022JointTask}.
The fatigue model of Xia would be relevant for musculoskeletal predictive simulation to provide optimal motions that minimize fatigue while driving the rigid body dynamics and muscle dynamics.

The three-compartment model ($3_{CC}$) has already been used alongside deep reinforcement learning to generate optimal motion of the upper limb~\citep{Cheema2020PredictingLearning}. 
To the best of our knowledge, the three-compartment model has not yet been used for optimal control problems. 
The three-compartment model requires that the sum of the three states, denoted as $\ma$, $\mr$, $\mf$, remains equal to 1 throughout the entire simulation. 
In direct single-shooting transcription, maintaining this equilibrium is straightforward. 
Only the discretized controls are the decision variables, and the states are derived from these controls and an initial guess at the starting point. 
Thus, we can choose an appropriate initial guess to satisfy the equilibrium.
However, this transcription is not appropriate for solving OCPs in biomechanics. 
When it comes to more efficient methods,   
such as direct collocation or direct multiple shooting~\citep{Dembia2019, Puchaud2023DirectSimulations}, two edge cases may arise when using the fatigue dynamics. 
The main reason is that both the discretized controls and states are decision variables, and continuity constraints connect the discretized states across successive intervals. 
First, the equilibrium may not be respected for each discretized state. 
Second, a drift can occur because of the tolerance of the continuity constraints.  
If the deviation of the fatigue state $\mf$ is smaller than what the continuity constraints allow, the OCP can find solutions where the fatigue state $\mf$ is increased in the next interval, preventing the model from getting fatigued.  
Adding an equality constraint in the OCP at each interval may seem straightforward, but it could potentially result in non-convergence issues because too many solutions can satisfy the equilibrium.  
As a result, we will adapt the three-compartment model for use in direct optimal control transcriptions, specifically, direct multiple shooting, in this study. 

Finally, from the experimental literature, we know that upper-limb neuro-muscular fatigue leads to modified muscle recruitments and kinematic strategies \citep{Tse2016AdaptationsFatigue, Cowley2017Inter-jointFatigue}.
While performing a repetitive pointing task, the range of motion for the trunk increased with fatigue, while that for the shoulder and elbow decreased~\citep{Bouffard2018SexMovements}.
Thus, it is expected that including fatigue dynamics would alter the initial motion after several repetitions of the task.
But, another issue related to predictive simulations is that a task should be repeated several times before fatigue can be observed. 
Repetitions may result in long-duration OCPs (exceeding 10 seconds).
Solving the predictive simulation for multiple repetitions would yield motion patterns that anticipate movements until the end of the task. 
However, individuals typically only anticipate the next few seconds or upcoming repetitions of a task and adapt their movements in response to the onset of fatigue.
Therefore, to observe the emergence of adaptive motion, we need to solve split OCPs as a sliding-horizon OCP.
Most biomechanical tasks, such as gait or reaching, can be studied cyclically. 
Thus, it would be relevant to solve a series of sliding-horizon OCPs, where the optimized window moves from cycle to cycle.

The first objective of Section~\ref{sec:fatigue_dynamics} is to present a modified $3_{CC}$ model that maintains the sum of the three compartments ($\ma$, $\mr$, $\mf$) equal to 1 in two edge cases that may occur in OCPs with direct transcriptions, such as direct multiple shooting.
The second objective of Section~\ref{sec:simulation} is to illustrate the relevance of the embedded fatigue dynamics of the modified $3_{CC}$ model in conjunction with an upper limb model to generate optimal solutions using a sliding-horizon OCP compared to a full-horizon OCP.
A dumbbell lifting task will be evaluated with three objectives, minimizing torque, fatigue, or torque and fatigue, respectively.

\section{Forward simulation of fatigue dynamics} \label{sec:fatigue_dynamics}
\subsection{Models and simulation method}

This section is divided into two subsections to highlight the added value of the modified fatigue model. 
After a brief reminder of the equations of the ($3_{CC}$) model, a forward simulation study will present the ability of the modified $3_{CC}$ model to maintain the equilibrium of Xia's states equal to 1 despite the two edge cases.

\subsubsection{The three-compartment and the stabilized three-compartment fatigue models} \label{sec:model_equations}

For the reader's convenience, the equations of the $3_{CC}$ fatigue model are reminded for one actuator.
The set of ordinary differential equations (ODE) with the active $\ma$, the fatigued $\mf$, and the resting $\mr$ states and one control known as the target load $\tl$ is:

\begin{equation}
\mdot= \begin{pmatrix}
\mrdot\\
\madot \\
\mfdot
\end{pmatrix}
= \begin{pmatrix}
-C(\ma, \mr, \tl) + r \times R \times \mf \\
C(\ma, \mr, \tl) -F \times \ma \\
F \times \ma - R \times \mf
\end{pmatrix}
= f_{3_{CC}}(m,\tl)
\end{equation}
where $R$ is the recovery rate, $F$ the fatigue rate, and $r$ the rest recovery multiplier introduced in the work of~\cite{Looft2018}. The feedback controller $C(\ma, \mr, \tl)$ is defined as:
\begin{subnumcases}{C(\ma, \mr, \tl) =}
L_D \times (\tl - \ma), & \text{if } $\ma < \tl$\text{ and } $\mr \geq  \tl - \ma$ \\
L_D \times \mr, & \text{if } $\ma < \tl$\text{ and } $\mr < \tl - \ma$ \label{eq:feedbackcontroller2}\\
L_R \times (\tl - \ma), & \text{if } $\ma \geq \tl$
\end{subnumcases}
where $L_D$ and $L_R$ are controller parameters defined for each actuator. 
This model can be applied to any type of actuator, such as muscle or joint torque.
For each actuator in a system, the three states are $\ma$, $\mr$, $\mf$, and the unique control is $\tl$.
To address the two edge cases presented in the introduction, we introduced an invariant stabilizer to the ODE, which was linked to $\mf$ state, in order to stabilize the three-compartment model ($3_{CC}$) as recommended in~\cite{michaud2022methodes}:

 \begin{equation}
 \begin{aligned}
\mfdot = F \times \ma - R \times \mf + S \; (1-\ma-\mr-\mf)
\end{aligned}
\end{equation}
with $S$ the stabilization coefficient. 
This added term stabilizes the fatigue dynamics by modifying the fatigue state, if the equilibrium ($\ma+\mr+\mf=1$) is not respected. 
This modified model and its associated ODE are denoted $3^{S}_{CC}$ and $f_{3^{S}_{CC}}$.

\subsubsection{Evaluation of the stabilized three-compartment model}

The Xia fatigue dynamics ($3_{CC}$) and its stabilized version ($3^{S}_{CC}$) were evaluated through two initial value problems summarized as study \#1, and \#2 in Table \ref{tab:test}.

\begin{table*}[!htbp]
\caption{Parameters of the two studies to evaluate the $3^{S}_{CC}$ fatigue model}
\centering  
\label{tab:test}
\begin{adjustbox}{max width=\textwidth}
\begin{threeparttable}
\begin{tabular}{ccc ccc cc}     
\toprule    
 \textbf{Study}	& \textbf{Fatigue models} & \textbf{Stabilizer coefficient} & \multicolumn{3}{c}{\textbf{Initial states}} & \textbf{Target load} & \textbf{Final time} \\
\# & & $S$ & $\ma$ & $\mr$ & $\mf$ & $\tl$ & (s) \\
\midrule       
\#1 & $3_{CC}$, $3^{S}_{CC}$ & 10 & 0.0 & 1.0 & 0.0 & 80\% & 60  \\
\midrule 
\#2 & $3^{S}_{CC}$  & 0, 5, 10, 20 &
\makecell{0.0 \\ 0.0} &
\makecell{1.0001 \\ 0.9999} &
\makecell{0.0 \\ 0.0} &
80\% & 60 \\
\bottomrule                                                          
\end{tabular}
\end{threeparttable}
\end{adjustbox}
\end{table*} 

\textbf{Simulations. }The study \#1 aimed to show the equivalence between the $3_{CC}$ and $3^{S}_{CC}$ models under the same initial states and target load. 
The initial states [$\ma$, $\mr$, $\mf$] were set to [0.0, 1.0, 0.0] (= 100\%) for both models. 
In study \#2, we evaluated the $3^{S}_{CC}$ model on the edge cases. 
In study \#2, two simulations were conducted. The $3^{S}_{CC}$ model had an initial sum of states with an error of $\pm 1e^{-4}$, coherent with the magnitude of errors encountered in OCPs due to continuity constraints.
The initial states [$\ma$, $\mr$, $\mf$] were set to [0.0, 1.0 $\pm$ $1e^{-4}$ , 0.0]. 
We also evaluated the effect of the stabilizer coefficient $S$~$\in$~\{0,~5,~10,~20\}.
In both studies, the initial value problems were solved with the Runge-Kutta 45 method using \textit{scipy} over $T = 60$~seconds with relative tolerance set to $1e^{-3}$ and absolute tolerance set to $1e^{-6}$. The target load was set to $\tl$ = 80\% of the maximum activation for the entire simulation time.
The parameters of the fatigue model used were taken from \cite{Looft2018} for an elbow torque actuator:
the fatigue and recovery rates $R$ and $F$ were both set to 0.00094 and 0.00912 and the controller parameters $L_D$ and $L_R$ were set to~10.

\textbf{Analyses. }For study \#1, the equivalence between $3_{CC}$ and $3^{S}_{CC}$ was measured through the root mean square error (RMSE) over the whole simulation for each state and their sum.
The absolute errors for each state $|\Delta\ma|$, $|\Delta\mr|$, $|\Delta\mf|$ at the final time (60 s) were also reported between the between $3_{CC}$ and $3^{S}_{CC}$. 
For study \#2, the simulation time for which the sum of states reached the precision orders from $1e^{-5}$ to $1e^{-9}$ was reported for each stabilization coefficient $S$.
The absolute errors $|\Delta\ma|$, $|\Delta\mr|$, $|\Delta\mf|$ at the final time (60 s) between the simulation starting from reference states $m=[0.0, 1.0, 0.0]$ and $m=[0.0, 1.0 \pm 1e^{-4}, 0.0]$ were also reported.

\subsection{Results of forward simulations}

\begin{figure}[!htbp] 
\includegraphics[width=0.8\linewidth, clip, trim={70 25 100 45}]{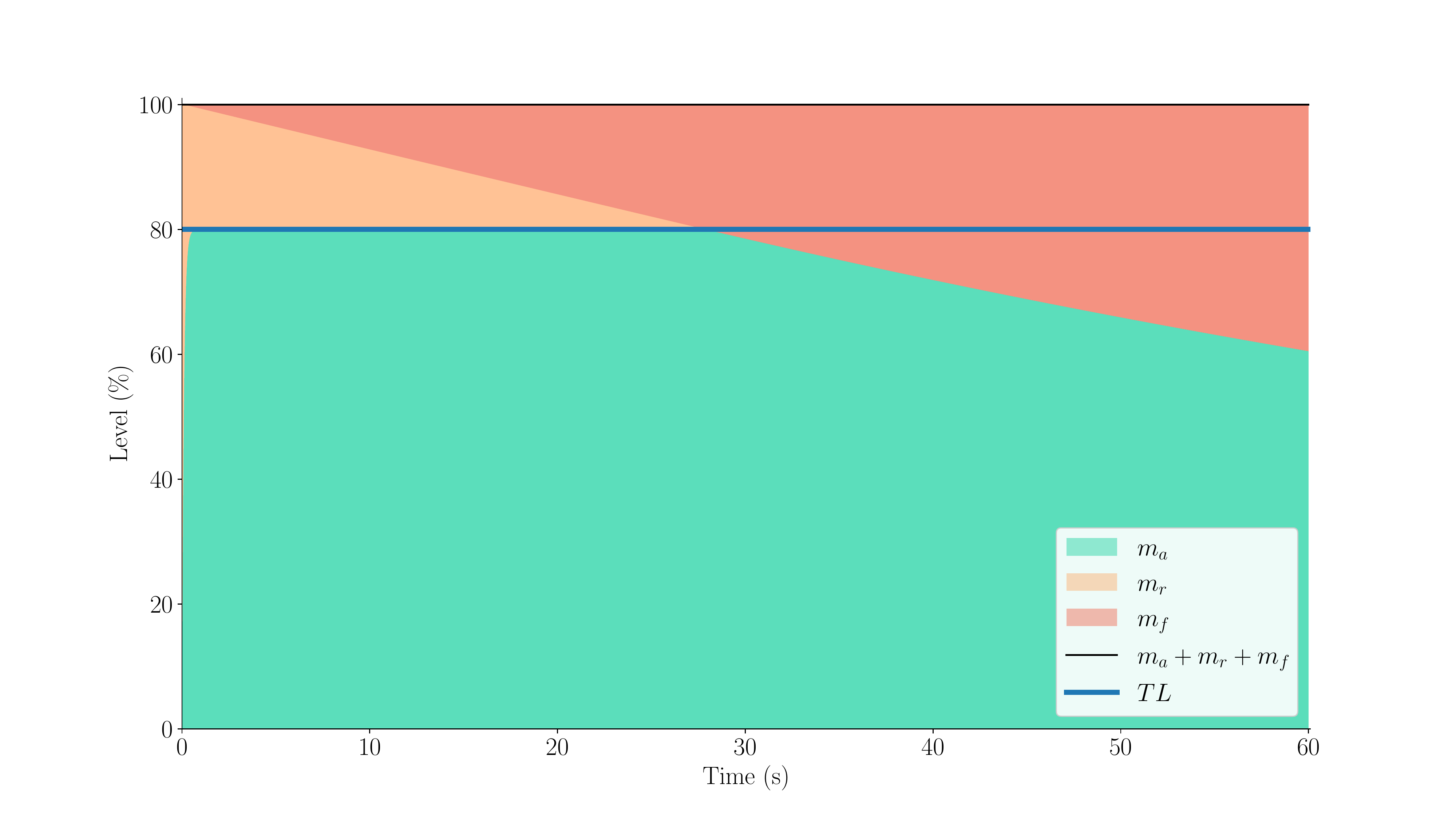}
\centering
\caption{The forward simulation of study \#1. The $3_{CC}$ and $3^{S}_{CC}$ models are superimposed. 
Active ($\ma$), fatigued ($\mf$), and resting ($\mr$) states for a 60-seconds simulation. 
$\tl$ stands for Target Load. }
\label{fig:study1}
\end{figure}

In study \#1, the $3_{CC}$ and $3^{S}_{CC}$ models showed equivalent numerical results over the whole forward simulation when starting with the same initial states (Fig.~\ref{fig:study1}).
The RMSEs for $\ma$, $\mr$, $\mf$ over the whole simulation were in the order of 1e\textsuperscript{-14}, and the differences for each state at $T = 60\si{s}$ were below 1e\textsuperscript{-14}.  
The difference between the two models is negligible.

In study \#2, the stabilizer coefficient $S = \{5, 10\}$ were the only stabilization coefficient that resulted in $ \ma + \mr + \mf \approx 1$ within a precision order inferior to 1e\textsuperscript{-5}. Coefficient $S=20$ ultimately diverged, failing to maintain stay within the initial order of precision of $\pm 1e^4$. For this coefficient, the sum of states $\ma + \mr + \mf$ kept oscillating around the objective while diverging. We only describe the results for $S = \{5, 10\}$ in the following, (Table~\ref{tab:study2}).
With $S = \{5, 10\}$, similar results were found for both simulations with $\mr=1 \pm 1e^4$.
The four simulations converged within a precision order of 1e\textsuperscript{-7} in 0.7 \si{s}. But only $S = 5$ converged below 1e\textsuperscript{-8}.
By the end of simulation time, the absolute errors $|\Delta\ma|$, $|\Delta\mr|$, $|\Delta\mf|$ between the simulations with the reference states and with the modified states were respectively below 1e\textsuperscript{-4}, 1e\textsuperscript{-6}, and 1e\textsuperscript{-4}.

\begin{table*}[!htbp]
\caption{Time in seconds for Study \#2 to reach different orders of precision on the sum of three pools for forward simulations of stabilized Xia model $3^{S}_{CC}$ with two different initial pools and stabilizer coefficients $S=\{5,10\}$. Absolute errors $|\Delta\ma|$, $|\Delta\mr|$ and $|\Delta\mf|$ for each state are also given.}
\centering  
\label{tab:study2}
\begin{adjustbox}{max width=\textwidth}
\begin{threeparttable}
\begin{tabularx}{\textwidth}{c *{15}{Y}}     
\toprule
&\multicolumn{3}{c}{\textbf{Initial states}} & \multicolumn{3}{c}{\textbf{Final states}}  
& \multicolumn{8}{c}{\textbf{Time (in s) to order}}   
\\
&\multicolumn{3}{c}{at 0 s} & \multicolumn{3}{c}{at 60 s} & \multicolumn{8}{c}{\textbf{of precision} satisfying}  \\
&\multicolumn{3}{c}{} & \multicolumn{3}{c}{} & \multicolumn{8}{c}{ $\ma + \mr + \mf \approx 1$}  \\
\cmidrule(lr){2-4} \cmidrule(lr){5-7} 
\cmidrule(l){8-15}
$S$ &$\ma$ & $\mr$ & $\mf$ 
& $|\Delta\ma|$ & $|\Delta\mr|$ & $|\Delta\mf|$
& \multicolumn{2}{c}{$1e^{-5}$} & \multicolumn{2}{c}{$1e^{-6}$} & \multicolumn{2}{c}{$1e^{-7}$} & \multicolumn{2}{c}{$1e^{-8}$}
 \\
\midrule       
5 & 0 & 1.0001 & 0 
& \mbox{7.04e\textsuperscript{-5}} & \mbox{2.57e\textsuperscript{-7}} & \mbox{7.06e\textsuperscript{-5}} 
& \multicolumn{2}{c}{0.46} & \multicolumn{2}{c}{0.92} & \multicolumn{2}{c}{1.4} & \multicolumn{2}{c}{1.89}
\\
5 & 0 & 0.9999 & 0 
& \mbox{7.06e\textsuperscript{-5}} &\mbox{4.67e\textsuperscript{-8}} & \mbox{7.06e\textsuperscript{-5}}
& \multicolumn{2}{c}{0.46} & \multicolumn{2}{c}{0.92} & \multicolumn{2}{c}{1.4} & \multicolumn{2}{c}{1.89}
 \\
\midrule       
10 & 0 & 1.0001 & 0 
& \mbox{7.07e\textsuperscript{-5}} & \mbox{7.21e\textsuperscript{-8}} & \mbox{7.05e\textsuperscript{-5}} 
& \multicolumn{2}{c}{0.23} & \multicolumn{2}{c}{0.46} & \multicolumn{2}{c}{0.70} & \multicolumn{2}{c}{-}
\\
10 & 0 & 0.9999 & 0 
& \mbox{7.04e\textsuperscript{-5}} &\mbox{2.56e\textsuperscript{-7}} & \mbox{7.06e\textsuperscript{-5}}
& \multicolumn{2}{c}{0.23} & \multicolumn{2}{c}{0.46} & \multicolumn{2}{c}{0.70} & \multicolumn{2}{c}{-}
 \\
\bottomrule                                                          
\end{tabularx}
\end{threeparttable}
\end{adjustbox}
\end{table*} 

To summarize, the stabilized model $3^{S}_{CC}$ was equivalent to the original $3_{CC}$ when the initial states were identical. The stabilization coefficients $S = \{5, 10\}$ were the only ones able to reach precision order below $1e^{-4}$. The stabilized model $3^{S}_{CC}$  reconciles the sum of the states to 1, increasing the precision order from $1e^{-4}$ to $1e^{-7}$ in 1.4 \si{s} for $S = 5$, and from $1e^{-4}$ to $1e^{-7}$ in 0.7 \si{s} for $S = 10$.

\section{Optimal control problem with embedded fatigue dynamics} \label{sec:simulation}
\subsection{Method for predictive simulations}

To highlight the relevance of embedded fatigue dynamics in OCPs, a fatigue-driven OCP with the stabilized three-compartment model $3^{S}_{CC}$ was built. The two degrees-of-freedom (DoF) ``arm26'' model \citep{Delp2007} was used to perform maximum repetitions of a biceps curl with a 3.6-\si{kg} dumbbell over 1~\si{s} each (1 round trip). We describe in this section the complete dynamics of the system, the full-horizon OCP, and an original three-cycle sliding-horizon OCP that both simulate the dumbbell lifting task. 

The rigid body dynamics is driven by positive (flexion) and negative (extension) joint torques denoted $\btau^{+} \geq 0$ and $\btau^{-} \leq 0$. Considering the 2-DoF model, the fatigue dynamics were applied to the four torque actuators, driven by their respective positive and negative torque activation ratios: $\Tilde{\tau}_i^{+}$ and $\Tilde{\tau}_i^{-}$ of each joint $i$~\citep{Cheema2020PredictingLearning}, replacing the theoretical target load $\tl$ previously described:

\begin{align}
 \label{eq:torque_ratio}
\Tilde{\tau}_i^{+} & = \frac{\tau^{+}_i}{\tau_{i}^{max}}, \\
 \Tilde{\tau}_i^{-} & = \frac{\tau^{-}_i}{\tau_{i}^{min}},
\end{align}
where $\tau_{i}^{max} = 50 \;\text{Nm}$ and $\tau_{i}^{min}=-50 \; \text{Nm}$ are the upper and lower bounds of each joint actuator, for shoulder and elbow \citep{Danneskiold-Samse2009}. 
The four fatigue dynamics of all actuators, described in~Section~\ref{sec:model_equations} were all included in a common function:
\begin{equation} \label{eq:sysdyn1}
\begin{aligned}
\frac{d\bm{\mstate}}{dt} = f_{3^{S}_{CC}}(\bm{\mstate},\Tilde{\btau}),
\end{aligned}
\end{equation}
with the torque activation ratios $\Tilde{\btau} = [\Tilde{\btau}^{+} \;\Tilde{\btau}^{-}]^\top$. For the four actuators, the recovery and fatigue rates $R$ and $F$ were set to 0.00094 and 0.456. The fatigue rate was fine-tuned to induce fatigued in a limited amount of time. The controller parameters $L_D$ and $L_R$ were set to~10.
The forward dynamics $f_{FD}$ is based on the minimal coordinates formalism as follows:
\begin{equation} \label{eq:sysdyn2}
\begin{aligned}
\qddot = f_{FD}(\q,\qdot,\btau) = M(q)^{-1} (\btau  - N(\q,\qdot) ),
\end{aligned}
\end{equation}
with $\btau = \btau^{+} + \btau^{-}$, $M$ the mass matrix, and $N$ contains the non-linear effects and the gravity forces.

The presented dynamics were included in the OCP transcribed using a direct multiple shooting approach.
The decision variables of the OCP included the discretized control trajectories $\controls=[\controls_0 \ldots \controls_n \ldots \controls_{N-1}]$ of size $N$, where $N$ is the number of intervals. The discretized state trajectories $\states=[\states_0 \ldots \states_n \ldots \states_N]$ of size $N+1$. 
The controls of the OCP, $\controls= [\btau^{+} \; \btau^{-}]^\top$, were discretized using piecewise constant functions. 
The states~$\states = [\q \; \qdot \; \bm{\mstate}]^\top$ contained the generalized coordinates $\q$ and the velocities $\qdot$ of the rigid-body dynamics, and the active $\bma$, fatigued $\bmf$, and resting $\bmr$ states of the fatigue dynamics all gathered in $\bm{\mstate} = [\bma \; \bmf \; \bmr]^\top$.
The system dynamics $f$ is written as a 1\textsuperscript{st}-order ODE form $\statesdot=f(\states,\controls)$ such as:
\begin{equation} \label{eq:sysdyn3}
\begin{aligned}
\frac{d}{dt}
\begin{pmatrix}
\q \\
\qdot \\
\bmdot
\end{pmatrix}
=
\begin{pmatrix}
\qdot \\
f_{FD}(\q,\qdot,\btau) \\
f_{3_{CC}}(\bm{\mstate},\Tilde{\btau})
\end{pmatrix}
=
f(
\begin{pmatrix}
\q \\
\qdot \\
\bm{\mstate}
\end{pmatrix},
\begin{pmatrix} \btau^{+} \\ \btau^{-} \end{pmatrix}
).
\end{aligned}
\end{equation}

Each OCP was designed to achieve a predefined number of repetitions of biceps curl of 1 second each.
Elbow angle $q_1$ was constrained to vary from 15 to 150$^\circ$ for each cycle. 
Three predictive simulations were considered with three objective functions $\Phi^{\bmf,\tau}$, $\Phi^{\bmf}$ and $\Phi^{\tau}$ designed to enforce the biceps curl motion and minimize fatigue.
To ensure the rigid-body model lifts the dumbbell while minimizing the motion of the shoulder (i.e., $q_0=0$), each of the objective functions minimized the motion of the shoulder joint angle $q_0$ with a quadratic Lagrangian term. 
We will refer to this term as shoulder penalty.
The cost functions $\Phi^{\bmf,\tau}$, $\Phi^{\tau}$ included a term to minimize torques and $\Phi^{\bmf,\tau}$, $\Phi^{\bmf}$ included a quadratic term on the fatigue states $\bmf$. 
A regularization term was applied to the torque rate of change $\Delta\btau_n = \btau_n -  \btau_{n-1}$ for each frame~$n$.
It yields three objective functions:
\begin{align} \label{eq:obj}
\Phi^{\bmf,\tau}(\states,\controls) =&  \;  \int_{t_0}^{t_{f}} \omega_{q_0} \; q_0^2 + \omega_{ \Delta\btau} \; \Delta\btau^2 +  \omega_{f}\; \bmf^2 + \omega_{\tau} \;\btau^2 \; dt, \\
\Phi^{\bmf}(\states,\controls) =& \;  \int_{t_0}^{t_{f}} \omega_{q_0} \; q_0^2 +  \omega_{ \Delta\btau} \; \Delta\btau^2
+  \omega_{f}\; \bmf^2 \; dt, \\
\Phi^{\tau}(\states,\controls) =&  \;  \int_{t_0}^{t_{f}} \omega_{q_0} \; q_0^2 + \omega_{ \Delta\btau} \; \Delta\btau^2 + \omega_{\tau} \;\btau^2  \; dt, 
\end{align}
where weights $\omega_{q_0}$, $\omega_{ \Delta\btau}$, $\omega_{f}$, $\omega_{\tau}$,  were set to $10^5$, $0.1$, $10^3$, and 1, respectively.

A torque activation inequality constraint was added to prevent the rigid-body dynamics from doing unrealistic motions in accordance with the fatigue level. 
\begin{align} \label{eq:torque_limit}
0 \leq \Tilde{\btau} + \bmf \leq 1,
\end{align}
In other words, this constraint prevents the rigid-body dynamics to behave independently from the fatigue dynamics.
Finally, continuity constraints inherent to the direct multiple shooting transcription and control and state bounds were set.

\begin{figure}[!htbp] 
\includegraphics[width=1\linewidth]{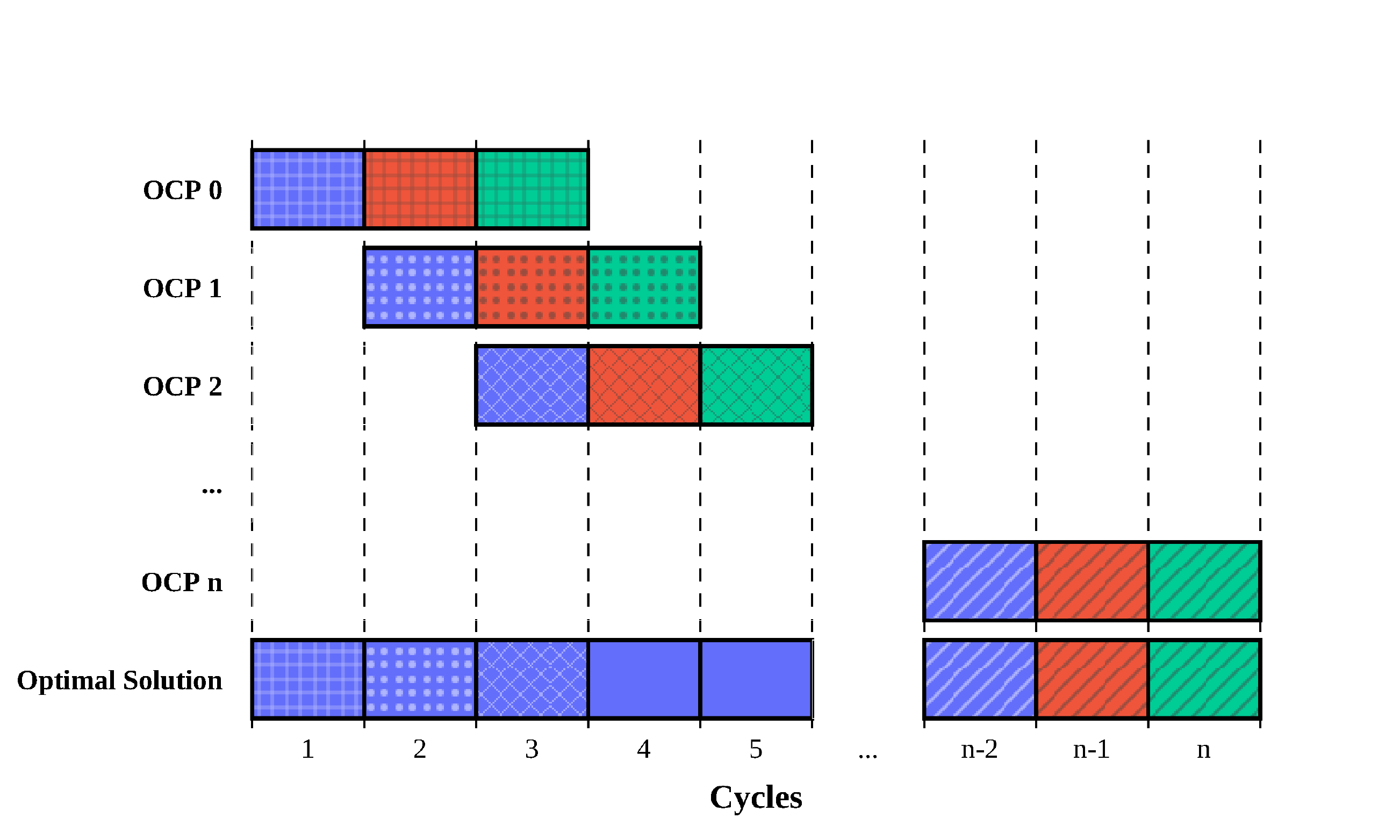}
\centering
\caption{Cyclic sliding-horizon OCP schematic. Each optimal control program (OCP) is a 3-cycle OCP. The moving window slides cycle by cycle. The stacked optimal solution comes from the 1\textsuperscript{st} cycle of each OCP. 1\textsuperscript{st}, 2\textsuperscript{nd}, and 3\textsuperscript{rd} cycles are respectively colored in violet, red and green. OCP 0, 1, and 2 are filled with squares, dots, and crosses respectively. Terminal OCP $n$ keeps the 2\textsuperscript{nd}, and 3\textsuperscript{rd} in the optimal solution. Initially introduced in the thesis of~\cite{michaud2022methodes}.}
\label{fig:nmpc}
\end{figure}

To evaluate the effect of the OCP formulation on the optimal solutions, a full-horizon, and a cyclic sliding-horizon OCP were considered to solve the maximum achievable dumbbell biceps repetitions. 
In total, six predictive simulations (cost functions $\times$ OCP formulations) were analyzed.
Full-horizon OCPs were solved for a predefined number of biceps curl repetitions. 
Each time an OCP was solved with success, another OCP with an extra repetition was solved  until convergence was no longer achieved.
We used this iterative approach to determine the maximum number of cycles, and only the penultimate OCP was considered for further analysis.
Initially introduced in the work of~\cite{michaud2022methodes}, the moving window of the cyclic sliding-horizon moved cycle by cycle and solved three cycles at once.
This method is often known as nonlinear model predictive control (NMPC), but it differs in that it operates on a cycle-by-cycle sliding-horizon, rather than focusing on the duration of the horizon.
The optimal solution of each  1\textsuperscript{st} cycle was kept as the final optimal solution, and the three cycles of the final window for which we obtained convergence, see Fig.~\ref{fig:nmpc}.
The initial guess of each OCP was based on the previous solutions.
As we slid the 3-cycle window, the initial guesses for the 1\textsuperscript{st} and 2\textsuperscript{nd} cycles retained the states from the 2\textsuperscript{nd} and 3\textsuperscript{rd} cycles of the previous OCP.
When initializing the 3\textsuperscript{rd} cycle of rigid-body states $[\q, \qdot]^\top$ and all the controls $\controls$, we duplicated the corresponding states and controls from the 2\textsuperscript{nd} cycle within the same window.
The states of Xia's model $[\ma, \mr, \mf]^\top$ were integrated from the duplicated states and controls to preserve the continuity of the fatigue dynamics between the 2\textsuperscript{nd} and the 3\textsuperscript{rd} cycle.

A fourth-order Runge-Kutta integrator (with five intermediate steps) was used to integrate the system dynamics, using Bioptim \citep{Michaud2022BioptimBiomechanics} on an Intel® Core™ i7-7920HQ CPU @3.10GHz with 32 Go RAM. The problem was solved using IPOPT with exact Hessian \citep{Wachter2006OnProgramming}, ma57 linear solver \citep{Duff2004MA57Systems}, and MX CasADi variables \citep{Andersson2019CasADi:Control}. Convergence tolerance and constraint tolerance of all presented OCPs were set to $10^{-6}$ and $10^{-4}$, respectively. The OCP was parallelized on eight threads. The code is available on the Github repository~\citep{github}.

\subsection{Analysis of predictive simulations}

We investigated the effects of cost functions ($\Phi^{\bmf,\tau}$, $\Phi^{\tau}$ and $\Phi^{\bmf}$) and OCP formulations (full-horizon \textit{vs} sliding-horizon) on: i) the maximum repetitions of dumbbell biceps curls; ii) CPU times to convergence; iii) cost functions value; iv) time series of Xia's model states.

\textbf{Cost Functions}. We analyzed the value of the three cost functions ($\Phi^{\bmf,\tau}$, $\Phi^{\tau}$ and $\Phi^{\bmf}$). More precisely, the task performance criterion minimized the shoulder joint angle. A higher shoulder angle indicates poorer task execution. The fatigue criterion minimized the fatigue states $\bmf$. The higher the fatigue states are, the less effective the actuator becomes because of Eq.~\ref{eq:torque_limit}. The value of these terms will be compared cycle to cycle and independently.

\textbf{Kinematic and Fatigue Analysis}. Elbow and shoulder angles $\q$ will be presented cycle by cycle for the six combined OCPs. We will emphasize the difference between the first and final cycles and compare the kinematics between them. The temporal series of fatigues states $[\bmr \; \bma \; \bmf]^\top$ will be presented.

\textbf{Stabilization effect}. To establish a connection with Section~\ref{sec:fatigue_dynamics}, we will outline the differences between the OCPs both with and without the invariant stabilizer at $S=10$. $S=10$ was picked because it was faster and accurate enough regarding the tolerance constraints.
This comparison will cover the maximum repetitions, the sum of pool accuracy, and the CPU time required for convergence.

\subsection{Results}

\begin{figure}[!htbp] 
\includegraphics[width=0.6\linewidth, trim={7 7 7 7}, clip]{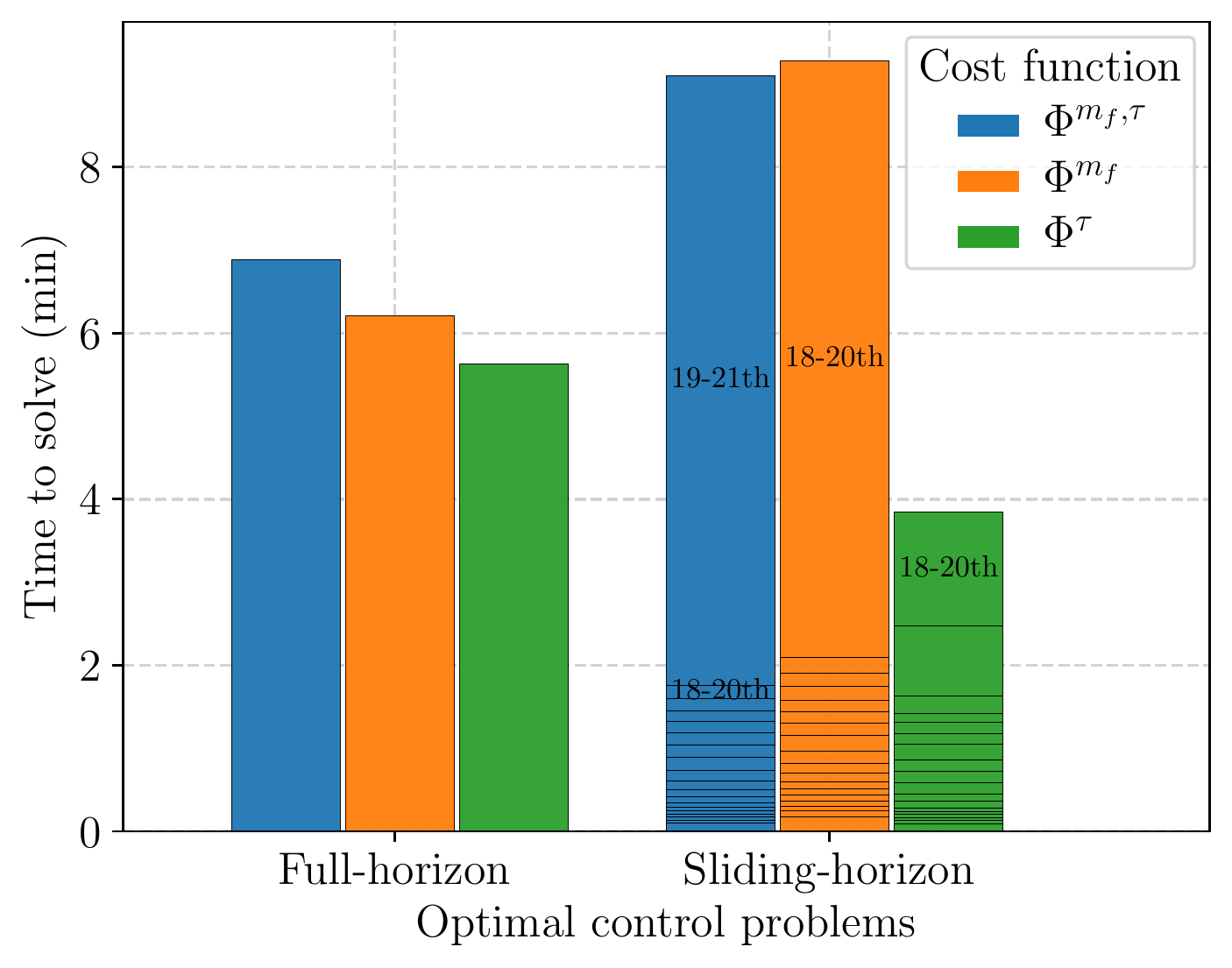}
\centering
\caption{CPU time to convergence for maximal repetition for full-horizon and three-cycle sliding-horizon OCP with the three considered cost functions $\Phi^{\bmf, \tau}$ (blue) $\Phi^{\bmf}$ (orange), and $\Phi^{\tau}$ (green). The CPU times of all sub-OCP for three-cycle sliding-horizon OCP are stacked together. The numbers of the terminal windows of sliding-horizon OCPs are displayed at the top of the stacked bar.}
\label{fig:time}
\end{figure}

\textbf{Convergence time}. The three full-horizon OCPs all converged until 32 cycles, and the three-cycle sliding-horizon OCPs with $\Phi^{\tau}$ and $\Phi^{\bmf}$ reached 20 cycles, and $\Phi^{\bmf, \tau}$ reached 21~cycles.
It is worth noting that the terminal windows of sliding-horizon OCPs took more time to solve. 
They are more challenging to solve because the solution space is reduced.
Full-horizon OCPs required more time to converge compared to sliding-horizon OCPs due to the inclusion of more cycles, as illustrated in Fig.~\ref{fig:time}. 
When normalized by the number of cycles, full-horizon OCPs still took more time to solve, ranging from 10.54 to 12.89~s/cycle, compared to the sliding-horizon OCPs, which ranged from 11.65 to~27.95~s/cycle. 
However, if we exclude the terminal windows, the sliding-horizon OCPs demonstrated faster solving times, ranging from 5.03 to~7.41~s/cycle.

\begin{figure}[!htbp] 
\includegraphics[width=1\linewidth]{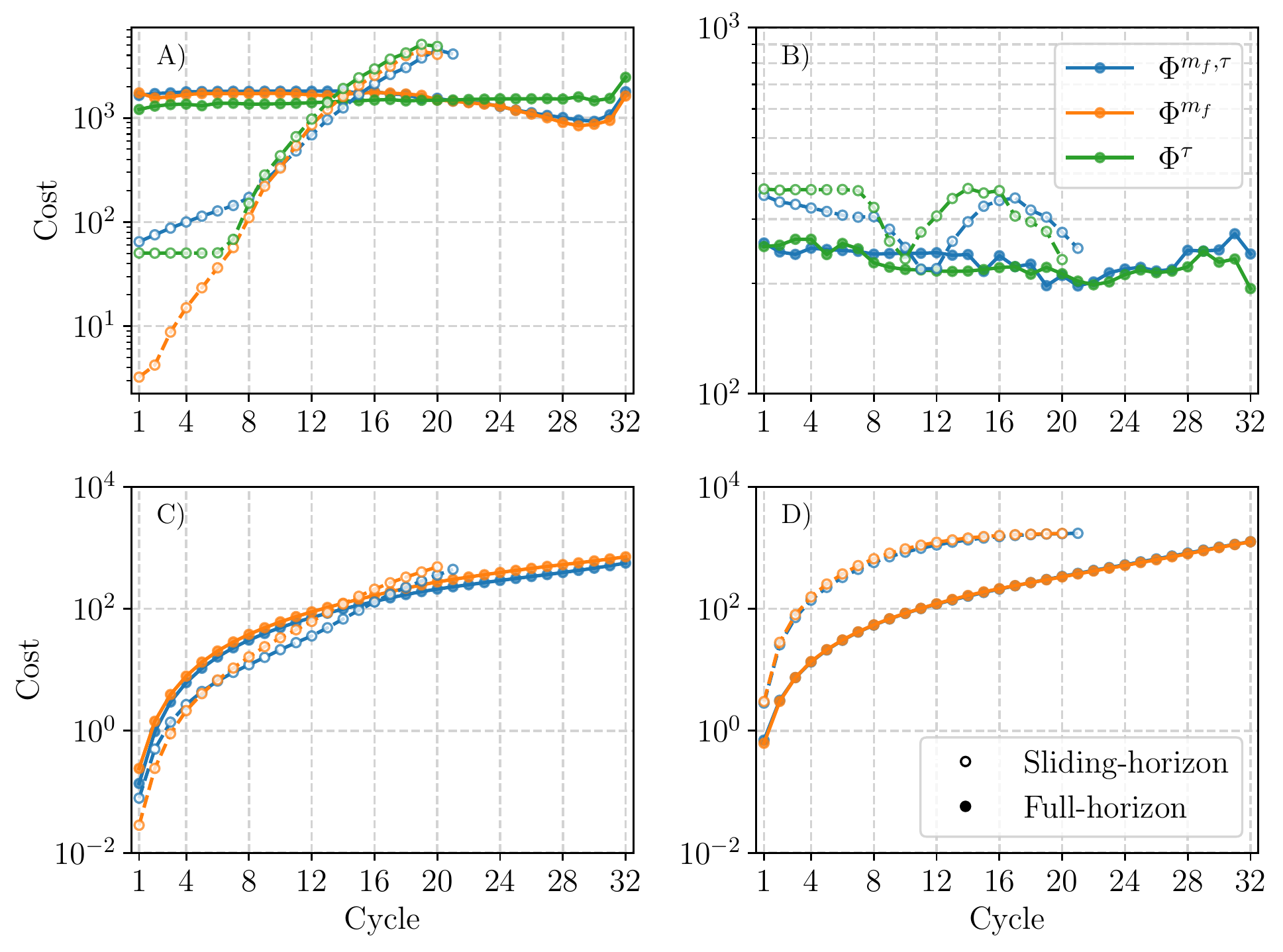}
\centering
\caption{Cost function terms evaluated for each cycle for full-horizon and three-cycle sliding-horizon OCP with the three considered cost functions $\Phi^{\bmf, \tau}$ (blue) $\Phi^{\bmf}$ (orange), and $\Phi^{\tau}$ (green): A) Lagrangian quadratic shoulder angle, B) Lagrangian quadratic torques, C) Lagrangian quadratic flexion fatigue states, D) Lagrangian quadratic extension fatigue states. Y-axes have logarithmic scales.}
\label{fig:cost}
\end{figure}

\textbf{Cost Functions}.
The shoulder penalty was lower for sliding-horizon OCPs from 1\textsuperscript{st} to 11\textsuperscript{th} cycle regardless of the cost functions compared to full-horizon OCPs (Fig.~\ref{fig:cost}.A). 
However, the flexion fatigue terms peaked for three-cycle sliding-horizon OCP, with values reaching 1718 and 1724 for $\Phi^{\bmf, \tau}$ and $\Phi^{\bmf}$ respectively, and 1252 and 1259 for their full-horizon OCPs equivalent. 
Interestingly, the antagonist extension fatigue terms were the highest for full-horizon OCPs until the 14\textsuperscript{th} and 17\textsuperscript{th} cycles. This suggests that extension torques were more actuated at the beginning of the task.

\textbf{Kinematic Analysis.} Larger differences were observed between the first and last cycles of the dumbbell biceps curls for three-cycle sliding-horizon OCP than for the full-horizon OCP; see Fig.~\ref{fig:kinematics} and the video provided as supplementary material. All cost functions increased at each cycle. The shoulder penalties reached their maximum at the final cycle because the model could not accomplish the task without moving the shoulder anymore. 
For the three-cycle sliding-horizon OCP, the kinematic strategy slowly deviated from an expected dumbbell lifting task to a swinging strategy, which aimed to lift the dumbbell as vertically as possible. In contrast, the full-horizon OCPs almost maintained the same optimal kinematic strategy throughout, except for the initial and final cycles, which differed from the other cycles as they were less constrained.

\begin{figure}[!htbp] 
\includegraphics[width=1\linewidth,trim={5 250 5 5}, clip]{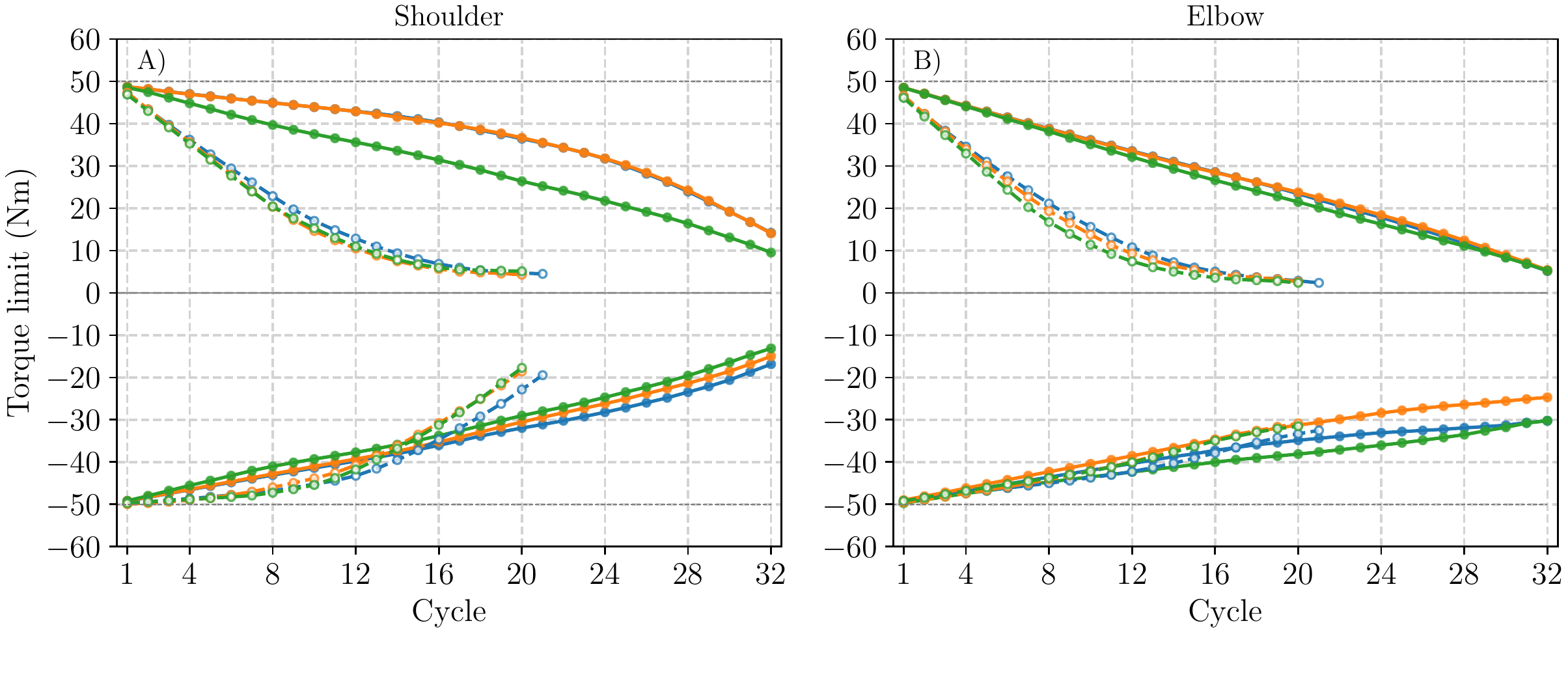}
\centering
\caption{Positive and negative torque limits at the end of each cycle computed from Eq. \ref{eq:torque_limit} for full-horizon (white dots) and sliding-horizon (colored dots) OCPs with the three cost functions $\Phi^{\bmf, \tau}$ (blue) $\Phi^{\bmf}$ (orange), and $\Phi^{\tau}$ (green): A) Shoulder torque limits, B) Elbow torque limits.}
\label{fig:tau_limits}
\end{figure}

\textbf{Fatigue Analysis.} 
The main reason for failure is the reduction of the maximum flexion torques of the shoulder and elbow, see Fig.~ \ref{fig:tau_limits}. When considering sliding-horizon OCP, the torque limits calculated using Eq.~\ref{eq:torque_limit} resulted in 10-fold and 16.5-fold reductions in maximum flexion torques of the shoulder and elbow, respectively. The maximal flexion torques decreased more gradually and more linearly for the full-horizon OCPs compared to the sliding-horizon OCPs. 
In addition, the maximal shoulder extension torques decreased earlier, indicating an earlier increased actuation of the shoulder in extension compared to the sliding-horizon OCP.

\begin{figure}[!htbp] 
\includegraphics[width=1\linewidth,trim={95 32 100 60}, clip]{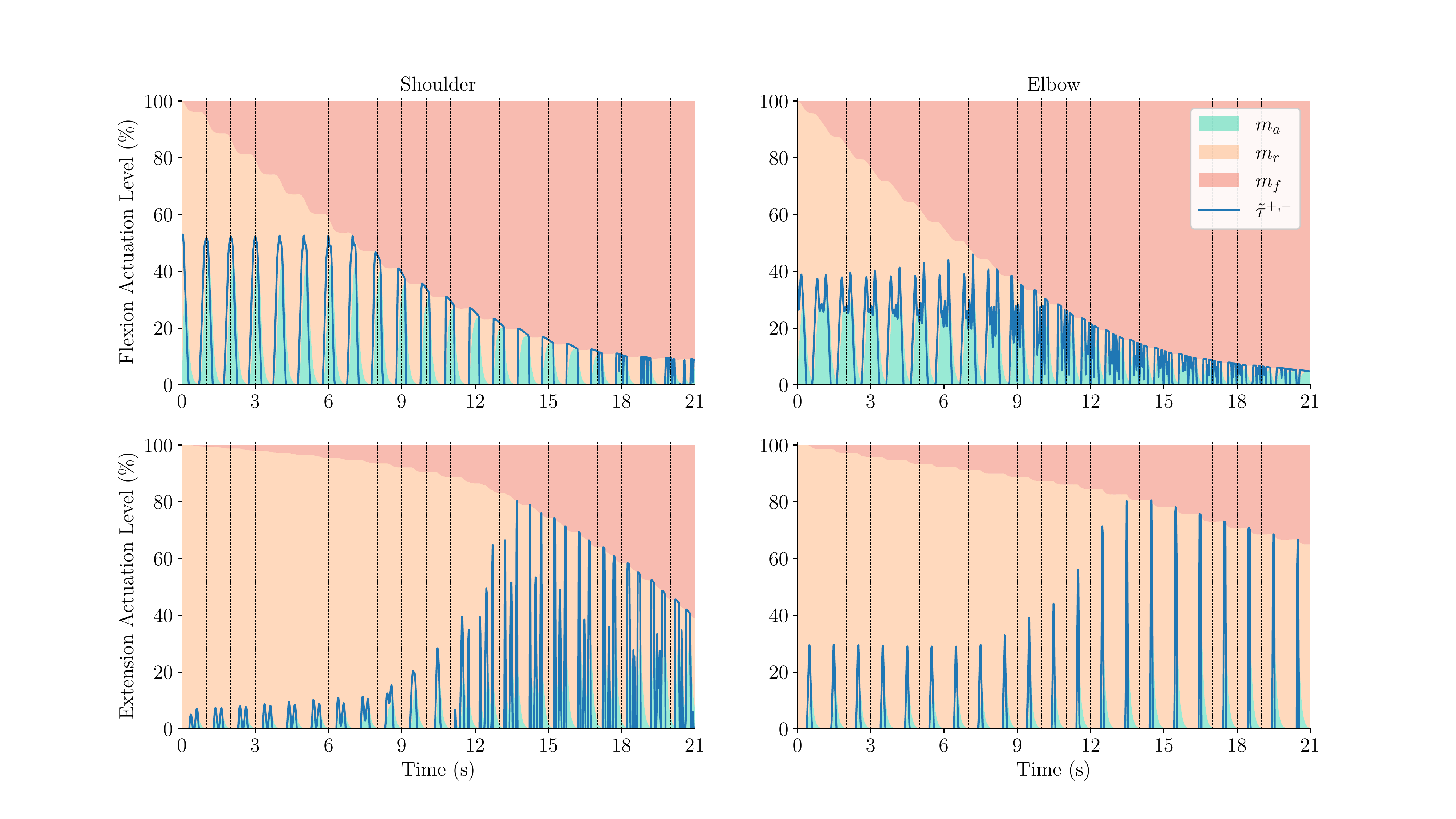}
\centering
\caption{
Stacked states of Xia's model, namely activation (green, $\ma$), rest (orange, $\mr$), and fatigue (red, $\mf$), for the four actuators, shoulder, and elbow in flexion and extension, and the torque activation levels (blue, $\Tilde{\btau}^{+}$ or $\Tilde{\btau}^{-}$) for sliding-horizon OCP with $\Phi^{\bmf,\tau}$.}
\label{fig:pools}
\end{figure}

Fig. \ref{fig:pools} displays the stacked states of Xia's model for the sliding-horizon OCP when using the cost function $\Phi^{\bmf, \tau}$. 
We can observe that the torque activation level more frequently reaches the upper bounds of Eq. \ref{eq:torque_limit} with each repetition of the task, indicating the onset of fatigue. 
For example, flexion shoulder and elbow torque activation levels reached upper bounds in cycle 7\textsuperscript{th} and 8\textsuperscript{th}, respectively. 
In extension, they were reached in the 13\textsuperscript{th} and 14\textsuperscript{th} cycles. 
Additional figures are available in Appendix~\ref{sec:annexe_pools}.

\textbf{Stabilization effect.}
Without the presence of the invariant stabilizer, the full-horizon OCPs were unable to match the maximum number of repetitions achieved when the invariant stabilizer was employed, see Table~\ref{tab:n_cycle_stabilizer}. 
For the three considered cost functions $\Phi^{\bmf, \tau}$, $\Phi^{\bmf}$, $\Phi^{\tau}$, they reached 20, 18 and 14 repetitions instead of 32. 
The absence of the invariant stabilizer did not affect the sliding-horizon OCPs for the cost functions that included the fatigue term 
$\Phi^{\bmf, \tau}$ and $\Phi^{\bmf}$. 
However, for the cost function $\Phi^{\tau}$, convergence was not achieved after 9 cycles. 
The maximum number of repetitions achieved without the invariant stabilizer were considered as references for further comparisons with OCPs that incorporated the invariant stabilizer.
When comparing the sum of pools, the invariant constraints were more precisely respected after convergence at the final time of the predictive simulations, see~Table~\ref{tab:sum_of_pools}. Also, the CPU time to convergence systematically decreased when the invariant stabilizer was utilized, see~Table~\ref{tab:cpu_time_stabilization}.

\begin{table*}[!htbp]
\caption{Number of cycles achieved for each OCP, with or without invariant stabilizer.}
\centering  
\label{tab:n_cycle_stabilizer}
\begin{adjustbox}{max width=\textwidth}
\begin{threeparttable}
\begin{tabular}{l ccc ccc}     
\toprule    
& \multicolumn{3}{c}{\textbf{Full Horizon}} & \multicolumn{3}{c}{\textbf{Sliding Horizon}}  \\
& $\Phi^{\bmf,\tau}$& $\Phi^{\bmf}$ & $\Phi^{\tau}$ & $\Phi^{\bmf,\tau}$ & $\Phi^{\bmf}$ & $\Phi^{\tau}$ \\
\midrule     
\# Cycles \textbf{with} invariant stabilizer & 32 & 32 & 32 & 21$^\star$ & 20$^\star$ & 20\\
\# Cycles \textbf{without} invariant stabilizer & 20 & 18 & 14 & 21$^\star$ & 20$^\star$ & 9\\
\bottomrule                                                         
\end{tabular}
\begin{tablenotes}
\item Notes: $^\star$ indicates that the same number of cycles has been achieved with and without invariant stabilizer
\end{tablenotes}
\end{threeparttable}
\end{adjustbox}
\end{table*}

\begin{table*}[!htbp]
\caption{CPU Time to convergence in minutes for all OCPs considering the maximum number of cycles reached by OCPs without invariant stabilizer.}
\centering  
\label{tab:cpu_time_stabilization}
\begin{adjustbox}{max width=\textwidth}
\begin{threeparttable}
\begin{tabular}{l ccc ccc}     
\toprule    
& \multicolumn{3}{c}{\textbf{Full Horizon}} & \multicolumn{3}{c}{\textbf{Sliding Horizon}}  \\
& $\Phi^{\bmf,\tau}$& $\Phi^{\bmf}$ & $\Phi^{\tau}$ & $\Phi^{\bmf,\tau}$ & $\Phi^{\bmf}$ & $\Phi^{\tau}$ \\
\midrule 
\# Cycles considered & 20 & 18 & 14 & 21$^\star$ & 20$^\star$ & 9\\
\midrule     
CPU time \textbf{with} invariant stabilizer (min) & 3.79 & 2.06 & 1.06 & 8.72 & 8.89 & 0.33\\
CPU time \textbf{without} invariant stabilizer (min) & 41.8 & 11.7 & 1.70 & 10.0 & 9.9 & 0.39\\
\bottomrule                                              
\end{tabular}
\begin{tablenotes}
\item Notes: $^\star$ indicates that the same number of cycles has been achieved with and without invariant stabilizer
\end{tablenotes}
\end{threeparttable}
\end{adjustbox}
\end{table*}

\section{General discussion}
We aimed to perform predictive simulations of repetitive human performance, taking fatigue dynamics into account. 
We have implemented an innovative three-cycle sliding-horizon OCP method to assess how movement progressively adapts to fatigue.
We first presented a stabilized version of $3_{CC}$ Xia's fatigue model to solve OCPs. The OCPs incorporated both rigid-body dynamics and fatigue dynamics to determine the maximum repetition capacity for the dumbbell biceps task.  
Our main findings are that: (i) the stabilized $3^{S}_{CC}$ fatigue model is suitable for forward simulation with edge cases;
(ii) full-horizon OCPs and three-cycle sliding-horizon OCP led to different optimal strategies and maximum numbers of repetitions;
(iii) the cost functions affect the early stages of three-cycle sliding-horizon OCP to limit the shoulder motion or the appearance of fatigue.

It is noteworthy that we adapted fatigue parameters to sustain motion until task repetition was no longer possible. Currently, no fatigue parameter set exists for dynamics tasks. 
Such parameters have only been identified in the context of isometric tasks \citep{Looft2018, Looft2020, Rakshit2021}. 
Identifying such parameters remains a challenge for future studies.
Based on our results, we can still extensively discuss the benefits of fatigue dynamics for predictive simulations.

\subsection{Evaluation of the stabilized three-compartment model} 

Adding an invariant stabilizer term is a common strategy when dealing with the integration of complex differential-algebraic equations~\citep{Ascher1997StabilizationSystems} particularly due to holonomic constraints in rigid-body dynamics \citep{Baumgarte1972StabilizationSystems}. 
We introduced such an invariant stabilizer on the fatigue state to use the Xia fatigue model in forward dynamics and direct multiple shooting optimal control transcription.  
In fact, when the sum of the states is not exactly equal to 1, the original fatigue model may result in inconsistent fatigue dynamics, leading to the non-convergence of OCPs.  
The invariant stabilization hastened CPU time to convergence and prevented early optimization failure, especially with full-horizon OCPs. 

Modifying the fatigue state $\mf$ when the sum of states is not equal to 1 was the most consistent choice because it would poorly affect the behavior of active and resting states, $\ma$ and $\mr$.  
Modifying the active state $\ma$ would result in a dysfunction of the fatigue dynamics and modifying the resting state $\mr$ would result in an additional resource for the system to satisfy the target load for a longer time. One extreme edge case could, nevertheless, make the fatigue dynamics dysfunctional: if the fatigue state is set to $\mf=0$ and the sum of the active and resting states $\ma+\mr>1$, the fatigue state would turn negative. Thus, we still recommend initializing the fatigue dynamics states carefully for OCPs.  

Initialization of OCP with random values without \textit{a priori} knowledge has been proposed in a multi-start approach to find better optimums \citep{Bailly2021OptimalModel}. It could fall into this extreme edge case prevented above and lead to OCP non-convergence. Finally, a high stabilization coefficient ($S\geq20$) induced oscillations around 1 of the states’ sum ($\ma + \mr + \mf$). This divergent behavior is undesirable in OCPs solved through direct multiple shooting or direct collocation approaches as the continuity constraints might become harder to satisfy. 
We recommend analyzing the effect of the stabilization coefficient prior to designing OCPs. 
The stabilized Xia fatigue model presented most of the properties we were looking for in handling the edge cases which can occur during OCPs and prevented early non-convergence issues.  

\subsection{Task definition} \label{subsec:task}

As shown experimentally \cite{Gates2008TheMovements}, only the goal-relevant features of task performance are preserved when fatigue appears.  
When performing dumbbell biceps curls, common kinesiologist instructions are first to fully flex the elbow and second to maintain the arm as vertical as possible to specifically strengthen the biceps. But we usually observe that athletes may disregard the second instruction when fatigue appears.  
Individuals do trade-offs between the recommended motion and their possible variations, circumventing the instructions initially given.  

Hence, achieving an elbow flexion of 150$^\circ$ was the main constraint in the present study. 
To simulate the compensation due to fatigue during the task, we slacked the constraint on the shoulder angle, i.e., arm verticality, which was defined as a penalty in the cost function. 
This soft constraint was slowly violated in sliding-horizon OCPs, whereas in full-horizon OCPs, it was violated from the first repetition.
Considering soft constraints provided more flexibility, thereby facilitating the achievement of kinematically admissible solutions.
In models offering more degrees of freedom, we anticipate observing additional compensatory mechanisms (e.g., trunk extension to initiate movement) and, perhaps, alternating multiple strategies to accomplish the task as long as possible.
Thus, we recommend carefully choosing the constraints, main objective, and penalties that define a task for predictive simulation with fatigue dynamics to could reveal compensatory mechanisms. 

\subsection{Fatigue dynamics in predictive simulations} 

The Xia's fatigue model was previously used with reinforcement learning approaches \citep{Cheema2020PredictingLearning, Wannawas2021Neuromechanics-basedCycling} for predictive simulations. To our knowledge, the current study is the first to incorporate Xia's fatigue model into an optimal control problem. The fatigue dynamics of actuators were interfaced with the rigid-body dynamics through the torque activation level.
Although the model was originally designed to control the target load using neural input, previous studies based on an inverse dynamics approach \citep{Silva2011AnMovements, Barman2022JointTask} used muscle forces or torques as input instead of the original target load.  
Accordingly, an extra constraint (Eq.~\ref{eq:torque_limit}) was added for OCPs to avoid inconsistent dynamic behavior between rigid-body and fatigue dynamics. Indeed, the torque activation level controls cannot be higher than $m_a+m_r$ to avoid situations where the actuation intent is superior to the resources available. 
 
There is still work to be done to drive the rigid-body dynamics with the outputs of the fatigue dynamics. Indeed, the central nervous system controls the muscle forces that produce motion through neural commands. Therefore, adopting a more human-like computational flow would enhance our understanding of motion planning under the influence of fatigue dynamics. 
In our biceps curl example, the fatigue states have almost no time to recover, see~Fig.~\ref{fig:pools}, even if the model is designed to enable recovery contrary to other fatigue dynamics available in the literature such as \citep{Ma2009AValidation}.  
Thus, finding strategies to recover from previous actuation would be possible. This could be achieved by combining more redundant musculoskeletal models (more DoFs and actuators) with the $3^{S}_{CC}$ Xia's fatigue model. Furthermore, by incorporating muscle actuators, we would expect the simulation not only to distribute the fatigue across joints but also across muscles, enabling the task to be sustained longer.

\subsection{Full-horizon vs sliding-horizon optimal control problems}

While the full-horizon OCP can generate kinematic and control strategies for a predefined number of cycles, this approach might not accurately reflect the way humans produce movement.  
It produces strategies that optimize the actuation of all cycles simultaneously to fit the objectives and constraints of the dumbbell biceps task.  
Thus, the full-horizon OCP infers the optimal solution from the decision variable values, including those of the final cycles. The full-horizon OCPs act as sensory feedback of fatigue on a motion we have not performed yet; that is why we must consider a shorter window. 

The three-cycle sliding-horizon OCP \citep{michaud2022methodes}, also known as nonlinear model predictive control or NMPC, considers a small window to produce its motion.
Therefore, it is suboptimal as opposed to the full-horizon solution.  
The sliding-horizon approach has been previously considered to optimize walking strategies on uneven terrain \citep{Darici2022ATerrain}, in accordance with the anticipatory use of gaze in acquiring information that looks ahead only three seconds in advance for objects \citep{Mennie2007Look-aheadTasks}, or one and half seconds for walking \citep{Matthis2018GazeTerrain}. 
In our study, we do not consider any visuomotor strategies but neuromuscular fatigue through torque actuation, aiming to maximize task performance duration. We may reasonably suppose that the perception of musculoskeletal fatigue can be evaluated over a finite time horizon, as considered in the present study. 

In the present application, the sliding-horizon OCPs were slightly longer to solve than the full-horizon OCPs. But if we consider factors such as: i) the time required for all full-horizon OCPs with the increase to solve the number of cycles (from 1 to 33 corresponding to the infeasible condition); ii) longer motions ($\geq$10 seconds) because of longer rest between reps or different parameters values of the Xia’s model; iii) and models with more than two degrees of freedom, the sliding-horizon approach is definitively a more efficient method.
The sliding-horizon OCP necessitates solving the OCP just once to estimate the maximum repetition of the task. 
Furthermore, this was the only approach that revealed adaptive strategies, which aligns with the anticipated adaptability in human behavior.
Therefore, we recommend considering a sliding-horizon OCP to plan long-duration motion trajectories to study fatigue adaptations. 

\subsection{Emergence of reactive vs anticipatory strategies} 

Neuromuscular fatigue causes modified muscle recruitments and kinematic strategies during upper limb pointing tasks, as supported by the literature \citep{Tse2016AdaptationsFatigue, Cowley2017Inter-jointFatigue, Bouffard2018SexMovements}. Commonly, the proximal range of motion increases to compensate for a decrease in the distal one. These adaptations are driven, in the presented OCPs, by the decrease in maximal torque available and the cost function.  
However, the cost function had nearly no effect on full-horizon OCPs, as maximizing the number of cycles was prioritized over the cost function.  

Indeed, since the fatigue states $\bmf$ should remain as low as possible over the whole task to achieve convergence, the fatigue states $\bmf$ were minimized no matter if they were included in the cost function. 
In contrast, in the sliding-horizon approach, the cost functions acted as sensory feedback for fatigue, especially in the early stages. 
According to the cost function, the sliding-horizon approach revealed two behaviors: reactive and anticipatory strategies to fatigue. 
When the cost function only penalized torques, the solution did not deviate until the torque reached its upper boundary at the 7th cycle, leading to a reactive strategy to fatigue.  
In other words, the current strategy was kept until it became no longer feasible. 
When the cost function penalized the fatigue states, the optimal solution slowly deviated with the sliding window before the torque reached its upper boundary.  
An anticipatory strategy emerged, which is consistent with the centrally driven adjustments that occur early on, as noted in \cite{Strang2009Fatigue-inducedAdaptation} on fatigue-induced adaptation.  The anticipatory strategy of the cost function $\Phi^{\bmf,\tau}$ led to one more repetition of the task, which might be attributed to the combination of torque and fatigue. 

To better fit how individuals regulate their effort, we could adjust the weight of each term in the cost functions, 
the same way the central nervous system modulates the weight of feedback linking muscle sensors and motoneurons.
Feedback informs individuals through sensory re-afferents based on the current level of fatigue experienced~\citep{crenna1987forward,duysens1995gating,prochazka1989sensorimotor}. It would be challenging to get an experimental measurement of how much individuals consider this sensory feedback. But one could consider an inverse optimal control approach to identify these weights~\citep{lin2016human}. 
The adaptations made in response to fatigue and sensory feedback are crucial for preserving the global performance of a predefined task, introducing movement variability, as observed in a study conducted by~\cite{yang2018changes}, where fatigue was identified as the primary factor contributing to the emergence of kinematic variability. 

In the present study, we could reproduce a similar behavior.  
Specifically, we noted an increasing reliance on the shoulder joint to accomplish the desired elbow flexion. This observation aligns with the findings from the aforementioned study~of~\cite{Bouffard2018SexMovements}, mimicking the distal to proximal range of motion transfer. 
Based on our findings, we recommend minimizing the fatigue states to induce an anticipatory kinematic strategy in optimal control predictive simulation.  
However, this approach may not be suitable if the primary goal is to strictly respect all the task instructions. 

\subsection{Task Expertise} 

The present study provides a valuable opportunity for examining the relationship between predictive simulation, fatigue, and task expertise, viewed through the prism of computational biomechanics. 
If one considers a perfect expert in a specific task, such as dumbbell curl biceps, this expert should be able to achieve a predefined number of repetitions based on their internal representations of force capabilities without failing.  
Based on empirical evidence, it is suggested that prior to initiating movement, a motor plan is constructed, functioning as a predetermined set of muscle commands that enable the seamless execution of the entire movement, unaffected by peripheral feedback. This concept, known as feed-forward control (see review \cite{desmurget2000forward}), considers the use of sensory information prior to the execution of movement based on an internal representation derived from past experiences.  
Thus, considering this perfect expert, the failure threshold would presumably remain unreached just by the end of the task.  
In our case, this would correspond to the trajectories found using the full-horizon OCPs. 

It is worth noting that such a theoretical level of expertise might be unattainable. 
Individuals may also rely on peripheral feedback during the execution of the task \citep{alves2020preplanned}, adapting their movements during the whole task. 
For example, alterations in pacing strategy during exercise were reported due to changes in muscle activation \citep{tucker2009physiological} because of afferent feedback from the various physiological systems and previous experience. Novices are, consequently, more susceptible to fatigue and early failure if they maximize the short-term performance (i.e., the cost function).  
The sliding-horizon OCPs may better reflect this behavior as they optimize the motion incrementally, three cycles at a time. The sliding-horizon OCPs only foresee the projection of fatigue appearance on the next two repetitions of the task and adapt.  
In summary, we can hypothesize that the larger the horizon, the more expert the subject is. 

Finally, task objectives can vary in their objectives, including fixed time, fixed distance, predefined repetitions, or a focus on lasting as long as possible or accomplishing them as quickly as possible.
All these types of tasks require managing fatigue differently. 
The presented predictive simulation framework sets the basis for understanding the interplay between fatigue and task achievement. 

\section{Conclusion} 

We presented a stabilized version of $3_{CC}$ Xia’s fatigue model, which was combined with rigid-body dynamics, to simulate fatigue dynamics when solving the OCP for the maximum number of repetitions of a dumbbell biceps task. The stabilized model is adapted to solve OCPs with its invariant stabilizer set on the fatigue compartment and a torque activation inequality constraint to avoid dynamics inconsistency between rigid-body and fatigue dynamics. Nevertheless, we recommend choosing the stabilization coefficient and the initial states to guarantee good convergence properties. 

The task must be carefully defined using a set of objectives and constraints, enabling adjustments from a reference motion as fatigue appears.
The sliding-horizon OCP is suited to plan long-duration motion trajectories since it only considers the current decision variables of the window to determine the optimal solution. 
This formulation revealed reactive strategies when including only torques in the cost function and anticipatory strategies when including fatigue states. 

By simulating fatigue dynamics and deviations from a reference motion, the optimal control approach could identify the most effective strategies for managing and balancing fatigue across different actuators.
Overall, the proposed approach has the potential to be a valuable tool in optimizing performance and reducing fatigue-related injuries in a variety of fields.

\bibliographystyle{unsrtnat}
\bibliography{references.bib}  

\appendix
\newpage
\setcounter{figure}{0}
\setcounter{table}{0}

\section{Joint angles}
\begin{figure}[!htbp]
\includegraphics[width=1\linewidth,trim={15 54 50 50}, clip]{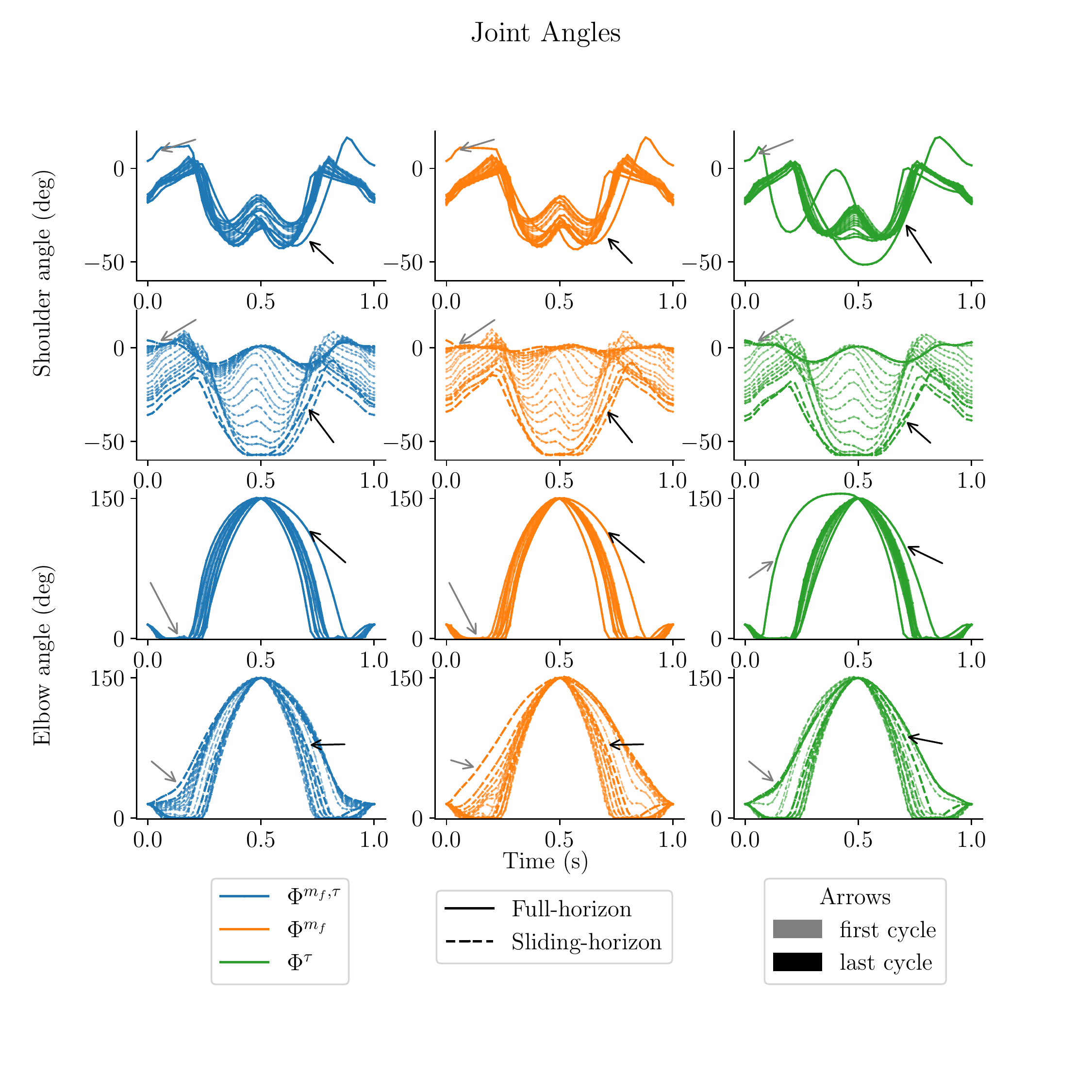}
\centering
\caption{Joint angles of shoulder and elbow are displayed for each cycle of full-horizon (solid line) and sliding-horizon (dashed line) OCPs, with the three cost functions $\Phi^{\bmf, \tau}$ (blue) $\Phi^{\bmf}$ (orange), and $\Phi^{\tau}$ (green). A gray and a black arrow designate the first and last cycle, respectively. This figure illustrates the joint angle deviations from the first cycle to the last cycle with the fatigue appearance.}
\label{fig:kinematics}
\end{figure}
\newpage
\section{Fatigue states for each OCP} \label{sec:annexe_pools}
\begin{figure}[!htbp]
\includegraphics[width=1\linewidth,trim={95 32 100 60}, clip]{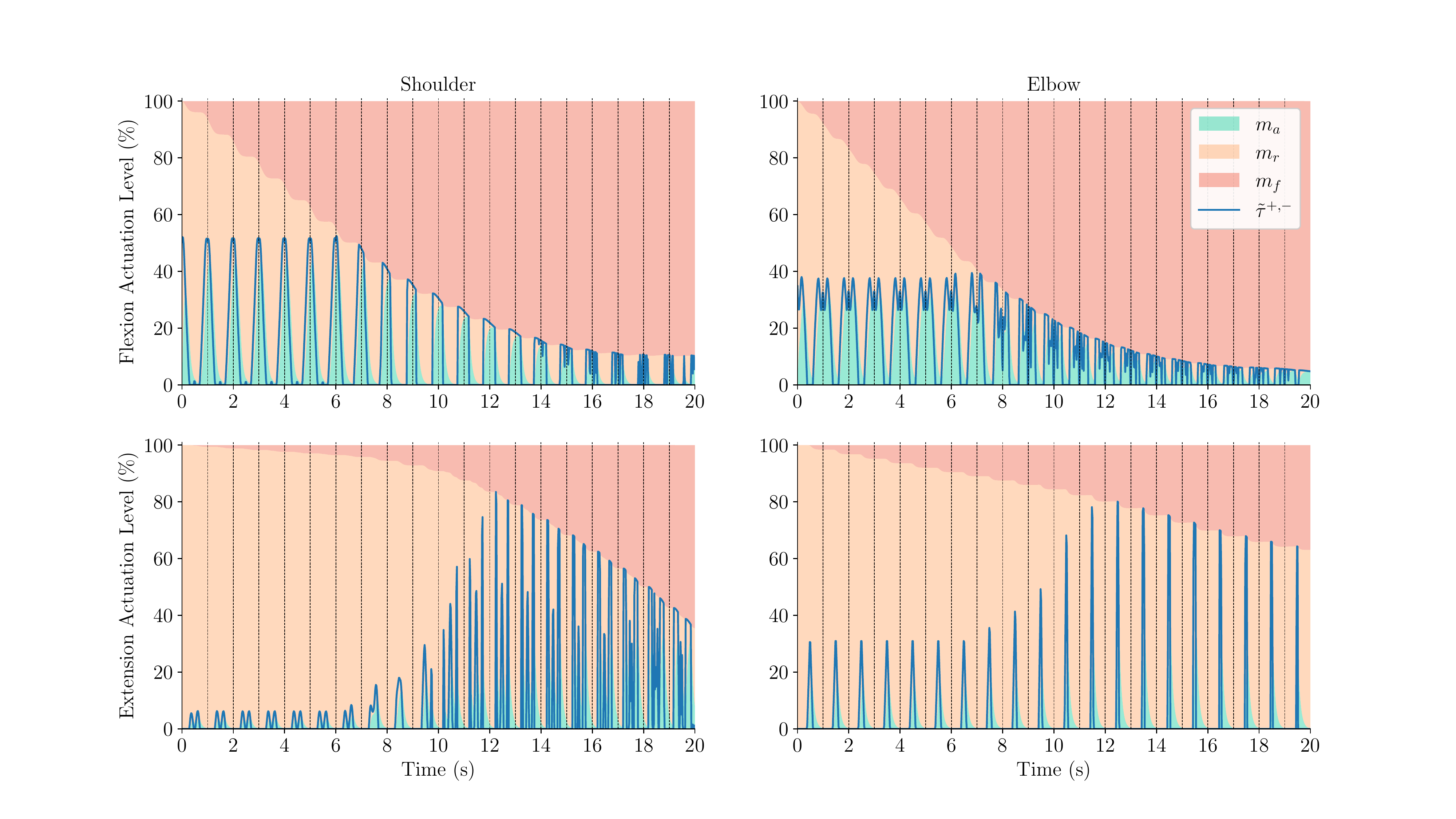}
\centering
\caption{
Stacked states of Xia's model, namely activation (green, $\ma$), rest (orange, $\mr$), and fatigue (red, $\mf$), for the four actuators, shoulder, and elbow in flexion and extension, and the torque activation levels (blue, $\Tilde{\btau}^{+}$ or $\Tilde{\btau}^{-}$) for sliding-horizon OCP with $\Phi^{\tau}$.}
\label{fig:pools_SH_tau}
\end{figure}
\begin{figure}[!htbp]
\includegraphics[width=1\linewidth,trim={95 32 100 60}, clip]{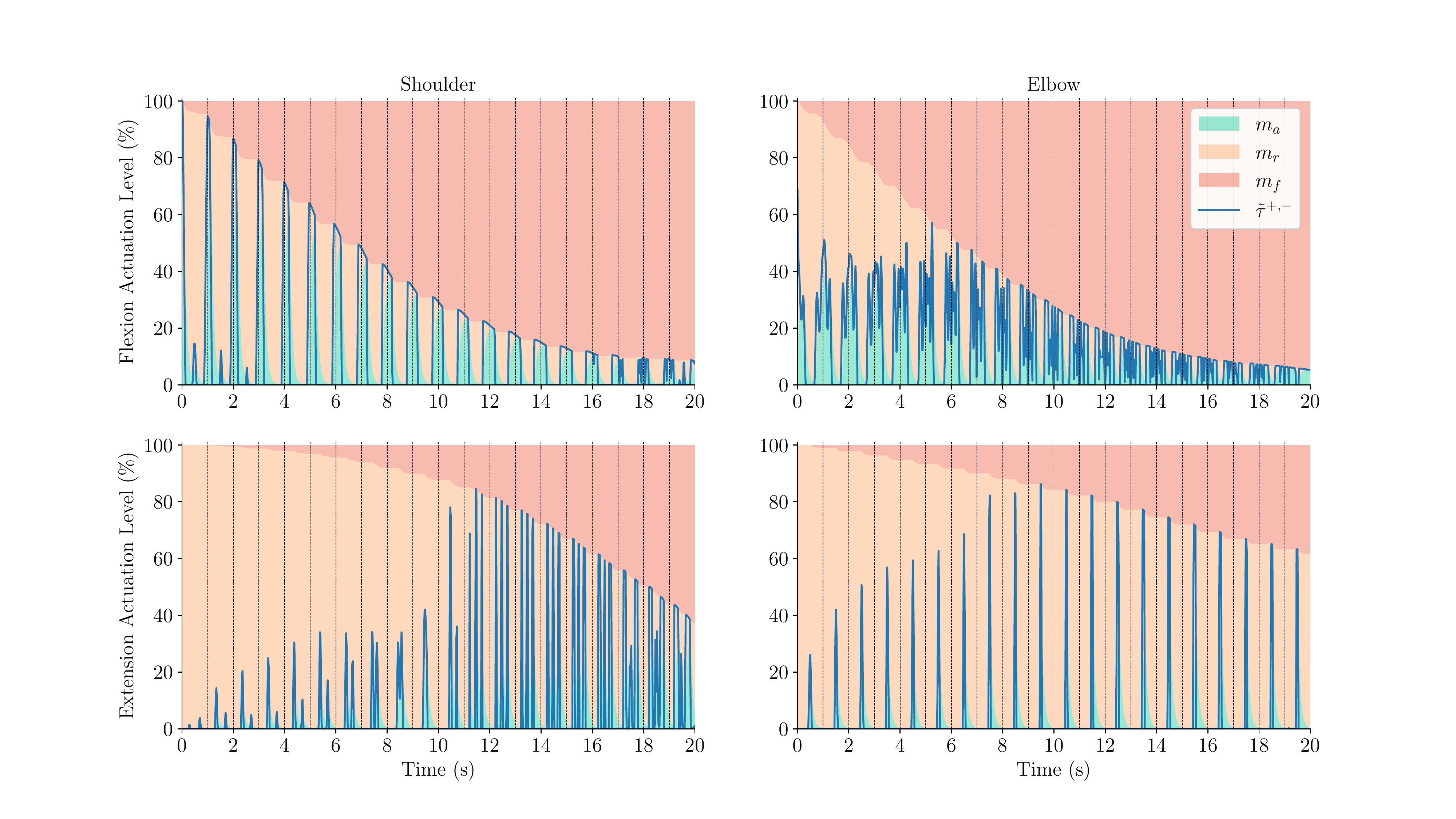}
\centering
\caption{
Stacked states of Xia's model, namely activation (green, $\ma$), rest (orange, $\mr$), and fatigue (red, $\mf$), for the four actuators, shoulder, and elbow in flexion and extension, and the torque activation levels (blue, $\Tilde{\btau}^{+}$ or $\Tilde{\btau}^{-}$) for sliding-horizon OCP with $\Phi^{\bmf}$.}
\label{fig:pools_SH_mf}
\end{figure}
\begin{figure}[!htbp]
\includegraphics[width=1\linewidth,trim={95 32 100 60}, clip]{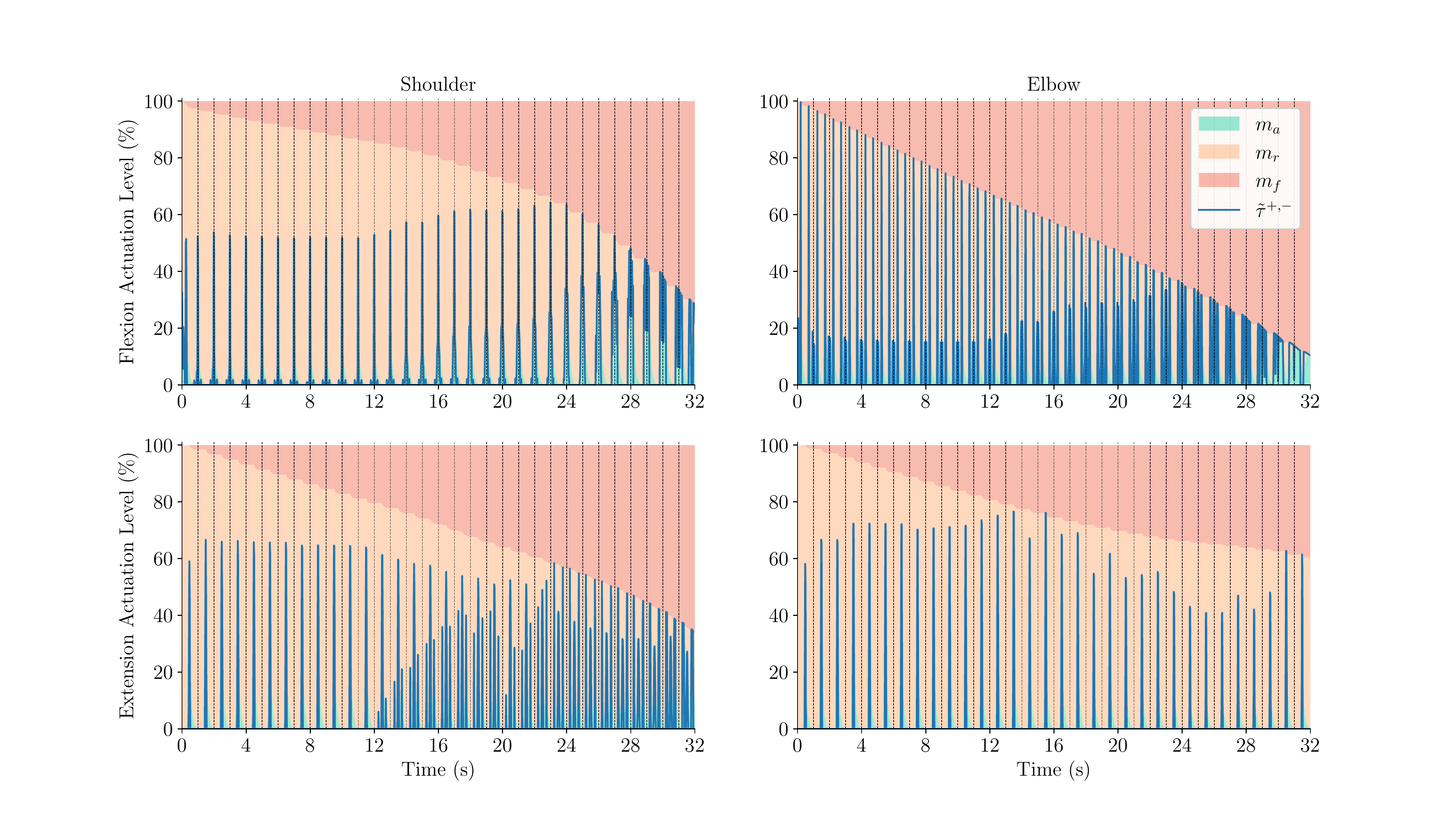}
\centering
\caption{
Stacked states of Xia's model, namely activation (green, $\ma$), rest (orange, $\mr$), and fatigue (red, $\mf$), for the four actuators, shoulder, and elbow in flexion and extension, and the torque activation levels (blue, $\Tilde{\btau}^{+}$ or $\Tilde{\btau}^{-}$) for full-horizon OCP with $\Phi^{\bmf, \tau}$.}
\label{fig:pools_FH_mf_tau}
\end{figure}
\begin{figure}[!htbp]
\includegraphics[width=1\linewidth,trim={95 32 100 60}, clip]{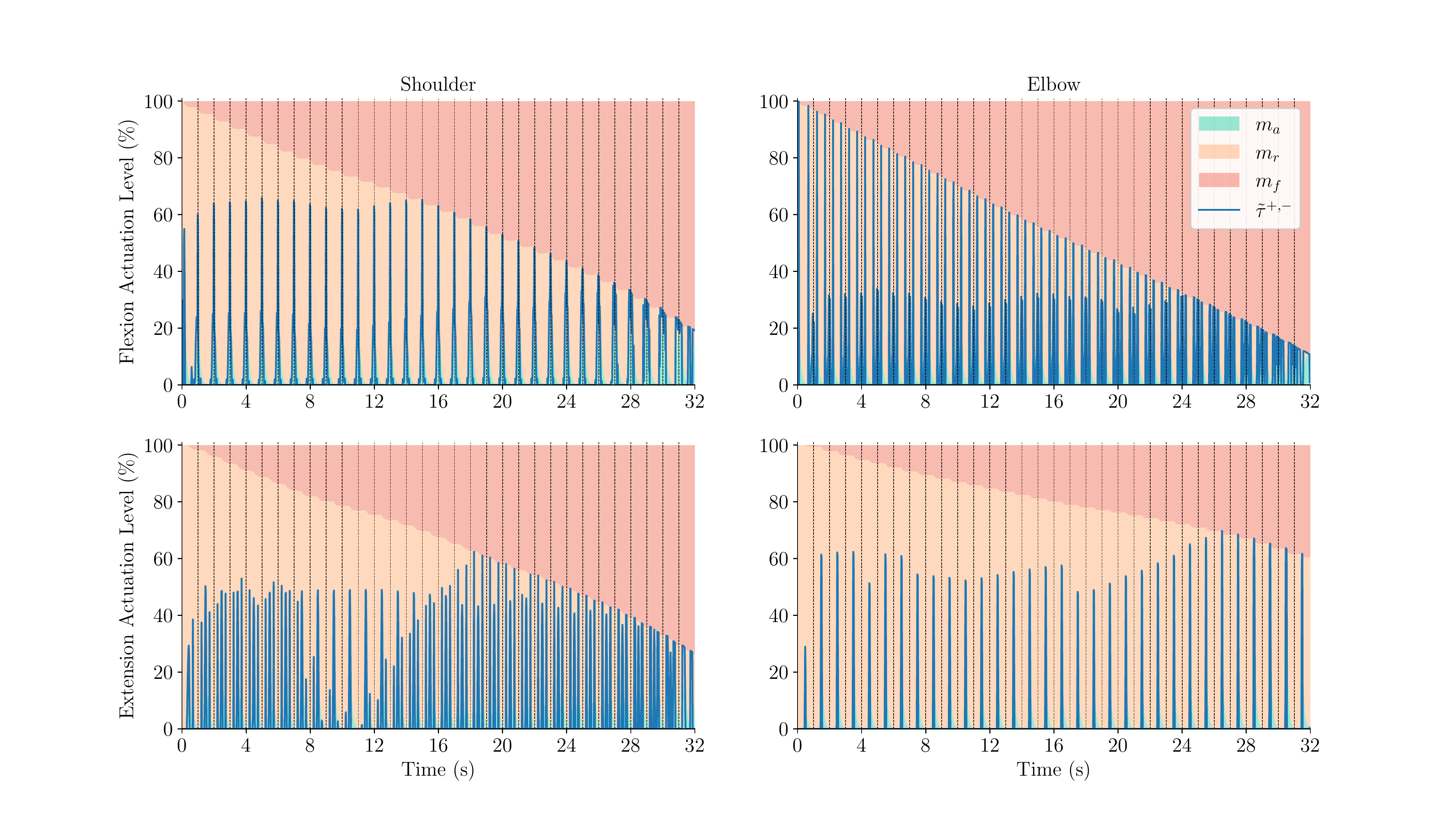}
\centering
\caption{
Stacked states of Xia's model, namely activation (green, $\ma$), rest (orange, $\mr$), and fatigue (red, $\mf$), for the four actuators, shoulder, and elbow in flexion and extension, and the torque activation levels (blue, $\Tilde{\btau}^{+}$ or $\Tilde{\btau}^{-}$) for full-horizon OCP with $\Phi^{\tau}$.}
\label{fig:pools_FH_tau}
\end{figure}
\begin{figure}[!htbp]
\includegraphics[width=1\linewidth,trim={95 32 100 60}, clip]{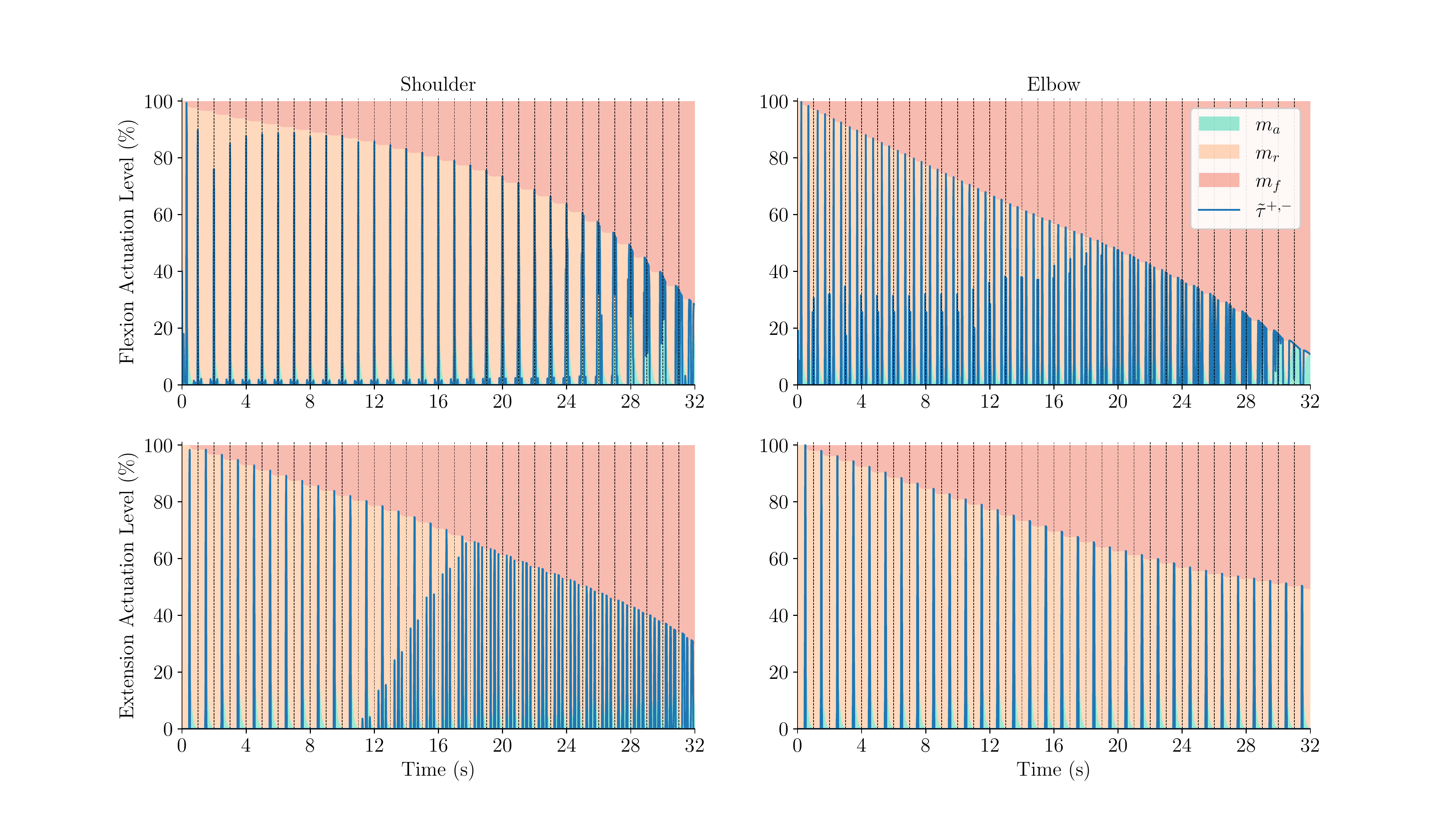}
\centering
\caption{
Stacked states of Xia's model, namely activation (green, $\ma$), rest (orange, $\mr$), and fatigue (red, $\mf$), for the four actuators, shoulder, and elbow in flexion and extension, and the torque activation levels (blue, $\Tilde{\btau}^{+}$ or $\Tilde{\btau}^{-}$) for full-horizon OCP with $\Phi^{\mf}$.}
\label{fig:pools_FH_mf}
\end{figure}

\newpage
\section{Evaluation of invariant constraint at the final time with and without invariant stabilizer.}

\begin{table*}[!htbp]
\caption{Invariant evaluation ($1 - (\ma + \mr + \mf) \approx 0$) at the final time for all OCPs considering the maximum number of cycles reached by OCPs without invariant stabilizer.}
\centering  
\label{tab:sum_of_pools}
\begin{adjustbox}{max width=\textwidth}
\begin{threeparttable}
\begin{tabular}{l ccc ccc}     
\toprule    
& \multicolumn{3}{c}{\textbf{Full Horizon}} & \multicolumn{3}{c}{\textbf{Sliding Horizon}}  \\
\midrule 
without invariant stabilizer & $\Phi^{\bmf,\tau}$& $\Phi^{\bmf}$ & $\Phi^{\tau}$ & $\Phi^{\bmf,\tau}$ & $\Phi^{\bmf}$ & $\Phi^{\tau}$ \\
\midrule 
Shoulder Flexion & $4.6\times10^{-6}$ & $2.4\times10^{-5}$ & $5.8\times10^{-9}$ & $1.3\times10^{-4}$ & $1.1\times10^{-5}$ & $2.3\times10^{-9}$ \\
Shoulder Extension & $1.5\times10^{-9}$ & $4.1\times10^{-5}$ & $2.2\times10^{-9}$ & $1.5\times10^{-3}$ & $1.2\times10^{-3}$ & $9.5\times10^{-12}$ \\
Elbow Flexion & $1.2\times10^{-8}$ & $1.3\times10^{-8}$ & $8.9\times10^{-9}$ & $1.4\times10^{-4}$ & $2.9\times10^{-4}$ & $8.1\times10^{-9}$ \\
Elbow Extension & $1.4\times10^{-8}$ & $1.3\times10^{-8}$ & $8.8\times10^{-9}$ & $2.6\times10^{-5}$ & $2.5\times10^{-7}$ & $7.7\times10^{-9}$ \\
\midrule 
with invariant stabilizer & $\Phi^{\bmf,\tau}$& $\Phi^{\bmf}$ & $\Phi^{\tau}$ & $\Phi^{\bmf,\tau}$ & $\Phi^{\bmf}$ & $\Phi^{\tau}$ \\
\midrule 
Shoulder Flexion & $2.4\times10^{-13}$ & $9.9\times10^{-12}$ & $5.0\times10^{-14}$ & $1.4\times10^{-8}$ & $8.8\times10^{-8}$ & $2.3\times10^{-9}$ \\
Shoulder Extension & $2.3\times10^{-12}$ & $2.3\times10^{-11}$ & $1.8\times10^{-13}$ & $9.3\times10^{-7}$ & $2.2\times10^{-6}$ & $9.5\times10^{-11}$ \\
Elbow Flexion & $3.7\times10^{-13}$ & $1.7\times10^{-12}$ & $3.2\times10^{-14}$ & $1.4\times10^{-7}$ & $3.0\times10^{-7}$ & $2.2\times10^{-16}$ \\
Elbow Extension & $2.0\times10^{-13}$ & $1.5\times10^{-13}$ & $1.1\times10^{-14}$ & $2.1\times10^{-9}$ & $1.2\times10^{-8}$ & $0.0\times10^{-20}$ \\
\midrule 
\# Cycles considered & 20 & 18 & 14 & 21 & 20 & 9\\
\bottomrule                                              
\end{tabular}
\end{threeparttable}
\end{adjustbox}
\end{table*} 

\end{document}